\documentclass[conference]{IEEEtran}
\IEEEoverridecommandlockouts
\usepackage{amsmath,amssymb,amsfonts}
\usepackage{algorithmic}
\usepackage{graphicx}
\usepackage{textcomp}
\usepackage{xcolor}
\usepackage[numbers,sort&compress]{natbib}
\usepackage{subfigure}
\usepackage[ruled]{algorithm2e}
\usepackage{stfloats}

\def\BibTeX{{\rm B\kern-.05em{\sc i\kern-.025em b}\kern-.08em
    T\kern-.1667em\lower.7ex\hbox{E}\kern-.125emX}}
\begin{document}
	

\title{A Fair Federated Learning Framework With Reinforcement Learning\\

 \author{\IEEEauthorblockN{Yaqi Sun$^{1,2 \ddagger}$, Shijing Si$^{2}$, Jianzong Wang$^{2*}$, Yuhan Dong$^{1*}$, Zhitao Zhu$^{2,3}$ and Jing Xiao$^{2}$}
 	\IEEEauthorblockA{$^{1}$Tsinghua Shenzhen International Graduate School, Shenzhen, China$^{\dagger}$\\
	\IEEEauthorblockA{$^{2}$Ping An Technology (Shenzhen) Co., Ltd., Shenzhen, China$^{\dagger}$\\
	\IEEEauthorblockA{$^{3}$Institude of Advanced Technology, University of Science and Technology of China, Hefei, China}  
		Email: yq-sun20@mails.tsinghua.edu.cn, shijing.si@outlook.com, jzwang@188.com,\\ dongyuhan@sz.tsinghua.edu.cn, andyzzt@mail.ustc.edu.cn, xiaojing661@pingan.com.cn}
	\thanks{$^{\ddagger}$ Work done as an intern at Ping An Technology (Shenzhen) Co., Ltd.
	\newline \hspace*{0.27cm}$^*$ Corresponding author: Jianzong Wang, \texttt{jzwang@188.com}, Yuhan Dong, \texttt{dongyuhan@sz.tsinghua.edu.cn}.
	\newline \hspace*{0.28cm}$^{\dagger}$ These units contributed equally to this work.
	}
	}}

}

\maketitle


\begin{abstract}
	Federated learning (FL) is a paradigm where many clients collaboratively train a model under the coordination of a central server, while keeping the training data locally stored. However, heterogeneous data distributions over different clients remain a challenge to mainstream FL algorithms, which may cause slow convergence, overall performance degradation and unfairness of performance across clients.
	To address these problems, in this study we propose a reinforcement learning framework, called PG-FFL, which automatically learns a policy to assign aggregation weights to clients.
	Additionally, we propose to utilize Gini coefficient as the measure of fairness for FL. More importantly, we apply the Gini coefficient and validation accuracy of clients in each communication round to construct a reward function for the reinforcement learning.
	Our PG-FFL is also compatible to many existing FL algorithms. We conduct extensive experiments over diverse datasets to verify the effectiveness of our framework.
	The experimental results show that our framework can outperform baseline methods in terms of overall performance, fairness and convergence speed. 
	
\end{abstract}


\section{Introduction}
Recent advances in machine learning techniques has enabled predictive algorithms to perform better than expected in certain domains. However, in real-world situations, a sufficiently effective model requires massive data for training. In some more sensitive scenarios, such as patient data from different hospitals, or driving data from different vehicles, a single may not have sufficient quantity and quality of data to learn a more robust model, and co-training may cause privacy leakage problems.\par

Federated learning (FL) is a new paradigm of distributed learning that aims to address the problem of communication efficiency in learning deep networks from decentralized data. Each client uses local data to learn local model parameters or parameter updates, then only transmits parameters to the server, and aggregates all parameters in the cloud, thereby obtaining a federated model without data exchange. However, practical applications show that heterogeneity causes non-trivial performance degradation in FL, including up to 9.2\% accuracy drop, 2.32 × lengthened training time, and undermined fairness \cite{Yang2021CharacterizingIO, Horvath2021FjORDFA}. For example, the data distribution of each client may be different, then the performance of the model, for example, between client A and client B, may vary greatly and the accuracy rate may even be lower than the prediction result of the local model.\par

\cite{Donahue2021ModelsharingGA, Blum2021OneFO} analyzed the willingness of clients to participate in the federal update in this case, and found that in some cases, clients with poor performance are more inclined to withdraw from the alliance. Based on this, referring to the fair methods for machine learning \cite{Cotter2019OptimizationWN, Dwork2012FairnessTA}, the entire network is faced with a choice, that is, the server hopes to maximize the global accuracy rate as much as possible, while each client hopes the final model to behave little differently than the other members. This problem can be described as a game theory problem of total cost allocation: the self-interested goal of all individual actors (fairness) and the overall goal of reducing total cost (optimality).\par

In addition, due to the different distribution of data on the clients, some clients with higher data quality may have more important predictive capabilities than others. In other words, the global may be overly dependent on the trained model of some clients. Therefore, when we start to pay attention to fairness and avoid this influence, intuitively, the convergence speed and prediction accuracy of the overall model may be affected \cite{Cui2021AddressingAD}.\par

Due to privacy issues that need to be guaranteed, we cannot directly access the raw data on each client, it is impossible to analyze the data distribution on the client. However, \cite{Wang2020OptimizingFL} demonstrates that there is an implicit connection between the distribution of training samples on the device and the parameters of the model trained based on these samples. To get as close as possible to the optimal solution, in this paper, we propose an extensible federated learning framework called Policy Gradient Fair Federated Learning (PG-FFL). PG-FFL can be regarded as an additional plug-in of the FL algorithm. Based on the policy gradient reinforcement learning algorithm, PG-FFL uses the local parameters of the client training model as observations, aims to balance the above problems of the model by assigning different aggregation weights to the clients participating in the update in each round of aggregation. The main contributions of this paper are as follows:\par
\begin{itemize}
	\item In this paper, we propose to utilize the Gini coefficient as the measure of fairness, which objectively and intuitively reflects the performance gap of the aggregated model among clients participating in federated training, and prevents polarization between client performance.
	
	\item We propose a fairness adjustment plug-in. For the federated model, we add the fairness indicator and use a plug-in based on the deep reinforcement learning (DRL) algorithm that can be used for any federated learning algorithm that does not involve aggregation weight adjustment. 
	
	\item In this paper, we port the policy gradient fair federated learning (PG-FFL) paradigm to two advanced FL optimization algorithms, namely FedAvg and FedProx. Experimental result shows that PG-FFL can significantly improve fairness in multiple datasets.
	
\end{itemize}

\section{Related work}
In this section, we introduce the main challenges federated learning faces and briefly introduce the current state of research.

\subsection{Federated Learning}
Federated learning is a distributed learning framework under differential privacy, which aims to learn a global model on the server-side using the model parameters learned by different local clients based on clients' private data \cite{McMahan2017CommunicationEfficientLO}.\par

In horizontal federated learning, the server trains a global model in specific aggregation ways iteratively aggregates local models from different clients. In each iteration, the server randomly selects a certain number of clients to transmit global model parameters, clients participate in the training using the downloaded global model for training, and then upload local training model parameters and aggregate the new global model on the server \cite{Wahab2021FederatedML, Li2020FederatedLC}.\par

\subsection{Fairness Challenges of Non-IID Data Distribution}

Classic federated learning algorithms aggregates the models of different participating clients by calculating a weighted average based on the amount of training data \cite{McMahan2017CommunicationEfficientLO}. However, in practical applications, the data and label distribution on different clients cannot fully meet the requirements for the distribution of all local client data and label IID distribution in the distributed algorithm. Therefore, the convergence and stability of federated learning are affected challenge \cite{Konecn2016FederatedLS, Karimireddy2019SCAFFOLDSC}. \cite{Xiao2021ANS, Yang2021HFLAH} proposes that part of the reason is the improper way of traditional federated learning’s server-side aggregation method.The contributions of clients in federated learning can be distinguished by their trained models’ validated accuracies.\par

Previous work has shown that non-IID data may bring parameter differences \cite{Zhao2018FederatedLW}, data distribution biases \cite{Hsieh2020TheND}, and unguaranteed convergence \cite{Sahu2020FederatedOI}, which can be improvemented both on the client side \cite{Sahu2020FederatedOI} and on the server side \cite{Huang2021BehaviorMD,Hsu2019MeasuringTE,Reddi2021AdaptiveFO}.\par


Due to the heterogeneity of data size and distribution on different clients in federated learning, simply aiming to minimize the total loss in large networks may disproportionately advantage or disadvantage the model performance on some of the clients, such as resulting in loss of uniformity of results across the clients \cite{Li2021DittoFA}. The accuracy of individual devices in the network, for example, cannot be guaranteed despite the high federated average accuracy.\par

There has been tremendous recent interest in developing fair methods for machine learning \cite{Cotter2019OptimizationWN, Dwork2012FairnessTA}, unfortunately, current methods can not apply to federated settings directly. The recent work introduces a fairness algorithm suitable for federated learning. \cite{Mohri2019AgnosticFL} uses a minimax optimization method to ensure that the overall fairness will not be improved at the expense of some client performance. \cite{Li2020FairRA} borrows the idea of resource allocation, fairness is allocated as a resource to achieve uniform distribution of clients’ performance, \cite{Wang2021FederatedLW} mitigates potential conflicts among clients before averaging their gradients. But these algorithms have fairness as the only goal, we can simply think that there is a permutation relationship between fairness and the best performance (usually expressed by the average performance). Therefore, in real federated learning applications, people will naturally want to further guarantee fairness when their programs can guarantee optimal performance.\par

Based on above, we propose a fairness adjustment plug-in. Our algorithm can be used for any federated learning algorithm that does not involve aggregation weight adjustment, and we add fairness considerations on the basis of pursuing the best performance.\par

\section{Fair Federated Learning} 
In this section, we first formally define the problem in Section A, then propose a naive solution in Section B combining deep reinforcement learning methods, and propose a new general framework in Section C, which can effectively handle the fairness disaster caused by non-iid data distribution while ensuring the performance of federated learning methods.

\subsection{Problem Statement}
The standard horizontal federated learning can be defined as to minimize\par

\begin{equation}
	\mathop{\min}_{\omega} f(x,\omega) = \sum_{i=1}^{N}p_{i}f_{i}(x,\omega),
\end{equation}\par

Where $f_{i}(x,\omega):=E_{x \sim P_i}[f{i}(x,\omega)]$, the vector $\omega$ denote model weights and $(x,y)$ denote a particular labeled sample, is the local loss function of the $i$-th client. Aggregation of loss functions of different clients, assuming that there are N clients partitioning data, where $D_i$ is the number of index sets of data points on client $i$, aggregation weight of the $i$-th client is defined as:\par
\begin{equation}
	p_i = \frac{D_i}{\sum_{i=1}^{N}D_{i}}.
\end{equation}\par

That is, simply think that the influence of a client on the global model is determined by its sample size. Training is a uniform distribution on the union of all samples, where all samples are uniformly weighted.\par

\begin{figure}[ht]
	\centering
	\subfigure[CIFAR-100 FedAvg IID]{\includegraphics[width=4.3cm, height=3.5cm]{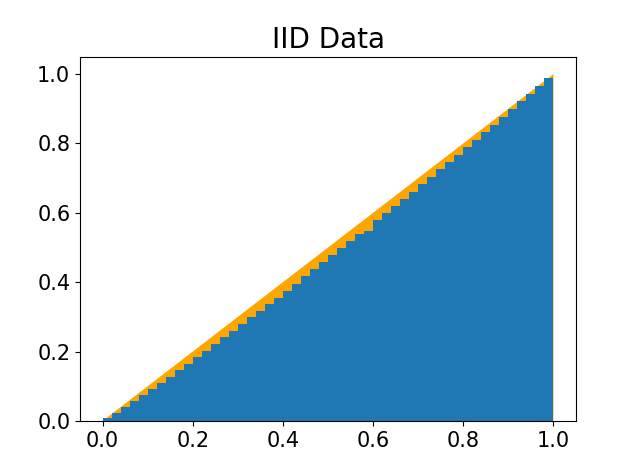}}
	\subfigure[CIFAR-100 FedAvg Non-IID]{\includegraphics[width=4.3cm, height=3.5cm]{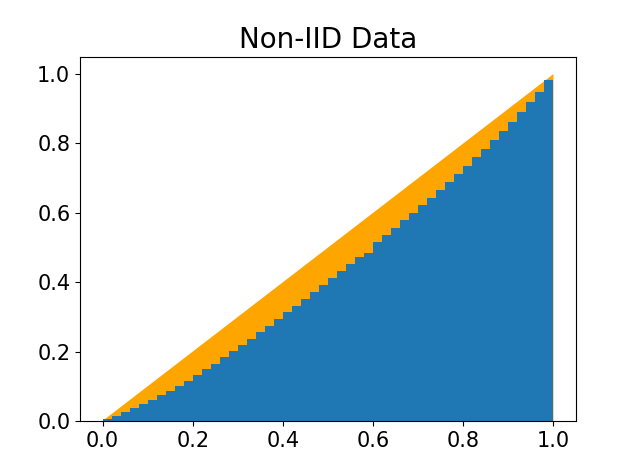}}
	\caption{When the CIFAR-100 data on 100 clients is in the IID distribution (left) and the non-IID distribution (right), the test results of FedAvg after 1000 rounds of training are shown in the figure. The larger the proportion of the orange takes, the more unfair it is.} 
\label{fig: IID_and_Non}
\end{figure}

The traditional federated learning method solves the problem of (1) jointly by (2) calculating the contributions of different clients, but this may cause the final global model to be biased towards clients with a large number of data points. Because of the considerable limitations in practical applications, we will not adopt this assumption in this paper, which we will illustrate through Fig.~\ref{fig: IID_and_Non}.\par

Considering some special cases in federated learning, such as joint training of different hospitals, different clients hope to jointly train a better model under the condition of protecting privacy, so the performance of different clients should not vary too much at this time. In order to reflect the fairness of a federated network, \cite{Wahab2021FederatedML, Li2020FederatedLC} proposed the uniformity and the standard deviation (STD) of testing accuracy between clients to measure network fairness. Unfortunately, indicators such as STD are related to the expectation of testing accuracy. Therefore, for different application scenarios, Assuming that all clients’ testing accuracy for networks A and B are 0.07, 0.08, 0.09 and 0.7, 0.8, 0.9, respectively, then A will have A smaller STD despite the "rich and poor" difference in test accuracy between the two networks. In order to alleviate this problem and make an index better measure the degree of network fairness, we put forward a new definition of fairness in the definition 1.\par

\textbf{Definition 1} (Fairness) \emph{We say a model $\omega_{1}$ provides a more fair solution than $\omega_{2}$ if the test performance of $\omega_{1}$ on $N$ devices, $\{acc_1,...,acc_N\}$, is more fair than that of $\omega_{2}$, i.e., $Gini\{F_{n}(\omega_{1})\}<Gini\{F_{n}(\omega_{2})\}$, where $F_{n}(\cdot)$ denotes the test accuracy on $N$ devices, and Gini\{·\} denotes the Gini Coefficient. Let $acc_{i}$ and $acc_{j}$ represents the accuracy on any client test set in the model, $\mu=\frac{1}{N}\sum_{n=1}^{N}acc_{n}$ represents the average accuracy of all clients.}

\begin{equation}
	Gini = \frac{\sum_{i=1}^{N} \sum_{j=1}^{N}|acc_i-acc_j|}{2 N^2 \mu}.
\end{equation}\par

We define the fairness of the model on all clients based on the Lorentz curve, and the indicator for judging the fairness is a proportional value between 0 and 1. The maximum Gini coefficient is 1 and the minimum is 0. The former indicates that the performance of the global model on the clients is absolutely uneven (that is, it performs best on one, and the rest are all 0), while the latter indicates that the global model is on the clients. The performance is absolutely average, and our goal is to ensure the final model performance while keeping Gini as small as possible.\par

Different from the existing FL system fairness definitions such as uniformity and STD, our proposed Gini describes the dispersion degree of constant distribution and has the characteristics of scale invariance. Therefore, the fairness degree of networks with different average performance can be compared with a uniform index.\par

Next, we prove through experiments that the non-IID data distribution on clients will not only reduce the accuracy and convergence efficiency but also lead to the disaster of fairness between clients. We trained the CIFAR-100 dataset with CNN model using FedAvg, set 100 clients and 10\% of the clients are selected to participate in the update in each round.\par

In the Lorenz curve, the smaller the proportion of orange, the higher the degree of fairness. As can be seen from Fig. 1, according to (2) updating the global model, the non-IID distribution of client data will increase the unfairness of the model's performance on the client side.\par

\subsection{DRL Settings}\label{AA}
In the federated learning training, since a certain percentage of clients are randomly selected to participate in the update in each round, the optimal weight distribution is non-differentiable. There are various approaches to deal with non-differentiable optimization bottlenecks, such as Gumbel-softmax \cite{Jang2017CategoricalRW} or stochastic back-propagation \cite{JimenezRezende2014StochasticBA}. In this paper, we model the problem of distributing the weight of different local models in the global model as a deep reinforcement learning problem to explore the optimal aggregation strategy \cite{Zhang2021DeepRL, Zhang2021AdaptiveCS}.\par

Instead of labeled data, reinforcement learning is a self-learning process in which agents maximize reward through interaction with the environment. The state-action transition and reward in the training process are abstracted as a Markov Decision Process(MDP), then the purpose of the DRL agent is to find an optimal policy that maximizes long-term reward expectations $\pi$:\par

\begin{equation}
	\pi^{*} = argmax_{\pi}E_{\tau \sim \pi(\tau)}[r(\tau)],
\end{equation}\par

where $\pi$ represents the policy, $\tau$ represents a trajectory obtained by using the policy to interact with the environment, and $r(\tau)$ represents the overall reward for this trajectory. Next, we expand the formula to obtain the objective function and the gradient based on the Monte Carlo approximation as: \par

\begin{equation}
	\ J(\theta) = E_{\tau \sim \pi_{\theta}(\tau)}[r(\tau)]
	=\int_{\tau \sim \pi_{\theta}(\tau)}\pi_{\theta}(\tau)r(\tau)d\tau,
\end{equation}\par

\begin{equation}
	\ \nabla_{\theta} J(\theta) = E_{\tau \sim \pi_{\theta}(\tau)}[\nabla_{\theta}log\pi_{\theta}(\tau)r(\tau)].
\end{equation}\par

Since our goal is to maximize long-term return expectations, we use gradient ascent to find the optimal policy \cite{Sutton1999PolicyGM}.\par

\begin{algorithm} 
	\caption{PGF-FedAvg} 
	\KwIn{Number of communication round $T$, number of clients $N$, percentage of updating in each round $C$, local epochs $E$, learning rate $\alpha$ and $\beta$.} 
	\textbf{Initialize:} Parameters $\phi^0$, $\omega^0$. \par
	\For{t=0,1,...,T-1} 
	{ 
		\textbf{Server} randomly selects a subset of clients $K=C*N$, and sends $\omega^{t}$ to them;\par
		\For{client $k \in [K]$ \textbf{in parallel}}
		{
			Client $k$ copies $\omega^{t}$ as local model parameters $\omega_{k}^{t}$;\par
			\For{j=0,1,...,E-1}
			{	Calculate gradient $g_k^j$\par
				$\omega_{k}^{j+1} \gets \omega_{k}^{j} - \alpha g_k^j $}
			$acc_k^t \gets$ \textbf{Accuracy}\{$\omega_{k}^{t}$ tests on validation dataset $\{ \mathcal{D}_k^V\}$\}
		}
		Calculate $\mu^t=\frac{1}{K}\sum^{K}_{k=1}acc_k^t$ and $Gini^t$ using (3)\par
		\textbf{DRL agent do:}\par
		Get reward $r^{t}=-\mu^t\log(Gini^t)$\par
		Update the DRL policy parameters $\phi$ \par
		\[\phi^{t+1} \gets \phi^{t}+\beta r^{t} \nabla_{\phi}log\pi(s_t,a_t)\]
		Get state $S^{t}=\{\omega_{1}^{t},...,\omega_{K}^{t}\}$\par
		Calculate Gaussian distribution mean $\{a_1^t,...,a_K^t\}$\par
		Calculate the aggregate weight $\{p_1^t,...,p_K^t\}$\par
		\textbf{DRL end}\par
		\textbf{Server} aggregates global model as
		\[\omega^{t+1} \gets \sum^{K}_{k=1} p_k \omega_{k}^{t}\]
	} 
\end{algorithm}

Therefore, the federated learning process can be modeled as a Markov Decision Process (MDP), where the state is represented by the model parameters of each client in each round. Given the current state, the reinforcement learning agent learns a policy distribution according to the policy calculate the aggregated weights corresponding to each client, thereby updating the global model. After that, the updated global model parameters are transmitted to the local. On the local validation set, the local validation accuracy of the global model will be observed, and the reward of the DRL agent will be obtained from the average validation accuracy and Gini coefficient of each client. function. The objective is to train the DRL agent to converge to the target accuracy and fairness level for federated learning as quickly as possible.\par

In addition, in our algorithm, the DRL agent only needs to obtain the local model parameters and validation accuracy on clients, neither introducing additional communication overhead, nor without collecting and checking any private information, which can achieve the purpose of privacy protection.\par

\textbf{State:} The state of the $t$-th round is represented by a vector $\{\omega_1^t,...,\omega_K^t\}$, which respectively represents the model parameters of K clients participating in the update. During the training process, the client and the server jointly maintain a list of model parameters $\{\omega_k^t {\mid}k{\in}K\}$. In each round of FL, the client updates the list after uploading the trained local model to the server.\par
\textbf{Action:} Each time the state list is updated, we use the client model parameters participating in the aggregation to train the DRL agent. The action space is composed of a vector $\{a_1^t,...,a_K^t\}$, thus the $k$-th client draws the aggregation weight $p^t_k$, where $p^t_k \in (0, 1)$, which comes from Gaussian distribution with a learnable mean $a_k^t$ and a unit variance, for participating in the aggregation of the global model.\par
\textbf{Reward:} The reward comes from a small verification set locally on the client, defined as $r^t=-\mu^t log(Gini^t$), where $\mu^t$ denotes the average validation accuracy on clients and $Gini^t$ denotes the fairness Gini coefficient of accuracy (see Definition 1), respectively. Such setting encourages federated models to achieve optimal and fair performance.\par

The DRL agent is trained to maximize long-term rewards based on $\gamma$ discounts:\par

\begin{equation}
	\ R = \sum_{a \sim \tau}\gamma^t r_t.
\end{equation}\par

Next, take FedAvg as an example to introduce our fairness optimization for the federated learning algorithm, the pseudo-code is shown in Algorithm 1.\par

\subsection{PG-FFL Workflow} 
We define federated learning on a classification problem. If there are $K$ clients in total, the training set, validation set and test set on the $k$-th client are respectively $D_k^{Train} = {(x_i,y_i)_{i=1}^N}{\sim}P_k$, $D_k^{Test} = {(x_i,y_i)_{i=1}^M}{\sim}P_k$ and $D_k^V = {(x_i,y_i)_{i=1}^L}{\sim}P_k$, where $x_i{\in}X_k $ is the d-dimensional feature vector of feature space $X_k$, and $y_i$ is its corresponding label. The training set, validation set, and test set on each client are independent and identically distributed (IID), but the data size and distribution on different clients are not required to be the same. The goal of our training is to learn a global model that performs well and is fair on the test set of each client.\par

\begin{figure}[ht]
	\centering
	\includegraphics[width=9cm, height=5.5cm]{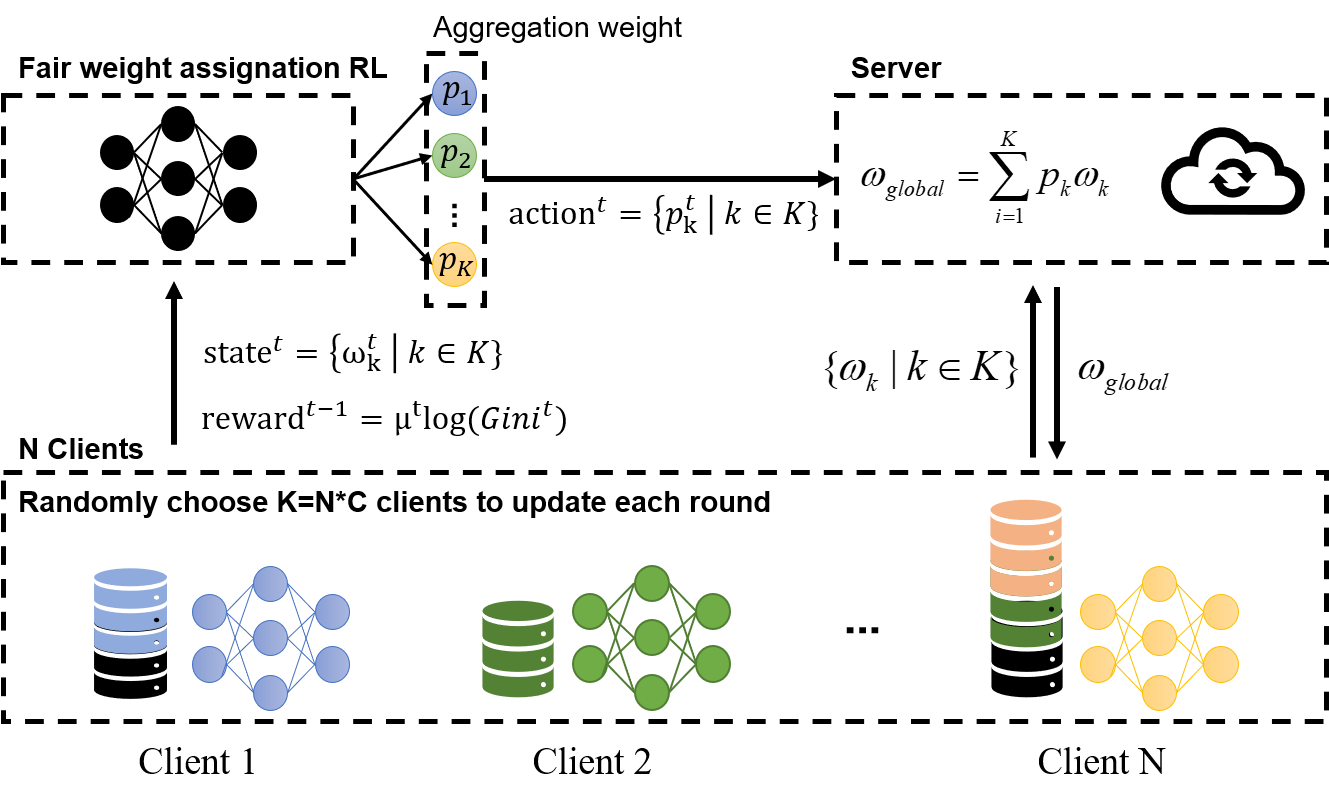}
	\caption{The framework of PG-FFL.}
	\label{fig: framework}
\end{figure}

Unlike traditional federated learning, our setup neither requires the data on different clients to be independently and identically distributed, nor is it designed to train a model that performs well on the server-side test set, which is more adapted to the requirements of the real world.\par

Fig.~\ref{fig: framework} shows how our algorithm, PG-FFL, is based on a reinforcement learning algorithm that assigns the aggregated weights of clients participating in the update in each round, following the steps below:\par%

\begin{itemize}
	\item \textbf{Step 1(initialization)}: All N available devices with non-identical data size and distribution check in server as clients, the server selects $K=N*C$ clients participating in the update according to a certain proportion C, initializes the model parameter $\omega^{init}$ and transmits it to the selected client, the client uses the global Model parameters get validation accuracy on the validation set, then train on local data, and return local model parameters $\{\omega_k^1, k\in K\}$ and validation accuracy $\{acc_k^1, k\in K\}$.
	\item \textbf{Step 2}: In the $t$-th round of iteration, the server calculates the average precision $\mu^t$ and Gini coefficient $Gini^t$ according to the returned $acc_k^t$, and then calculates the weight $P_{k}^t$ for the client k to participate in the global update, and then according to $\{\omega_k^t, k\in K\}$and $\{p_k^t, k\in K\}$ update the global model parameters.
	\item \textbf{Step 3}: The server randomly selects a certain percentage of clients to participate in the update. After the selected clients use the last round of global model $\omega^{t-1}$ for local training, upload the locally updated model parameters and verification accuracy.
\end{itemize}

\hspace{-1cm}
\begin{figure*}[bp]
	\centering
	\subfigure[Case \uppercase\expandafter{\romannumeral1}]{\includegraphics[width=6cm, height=4.3cm]{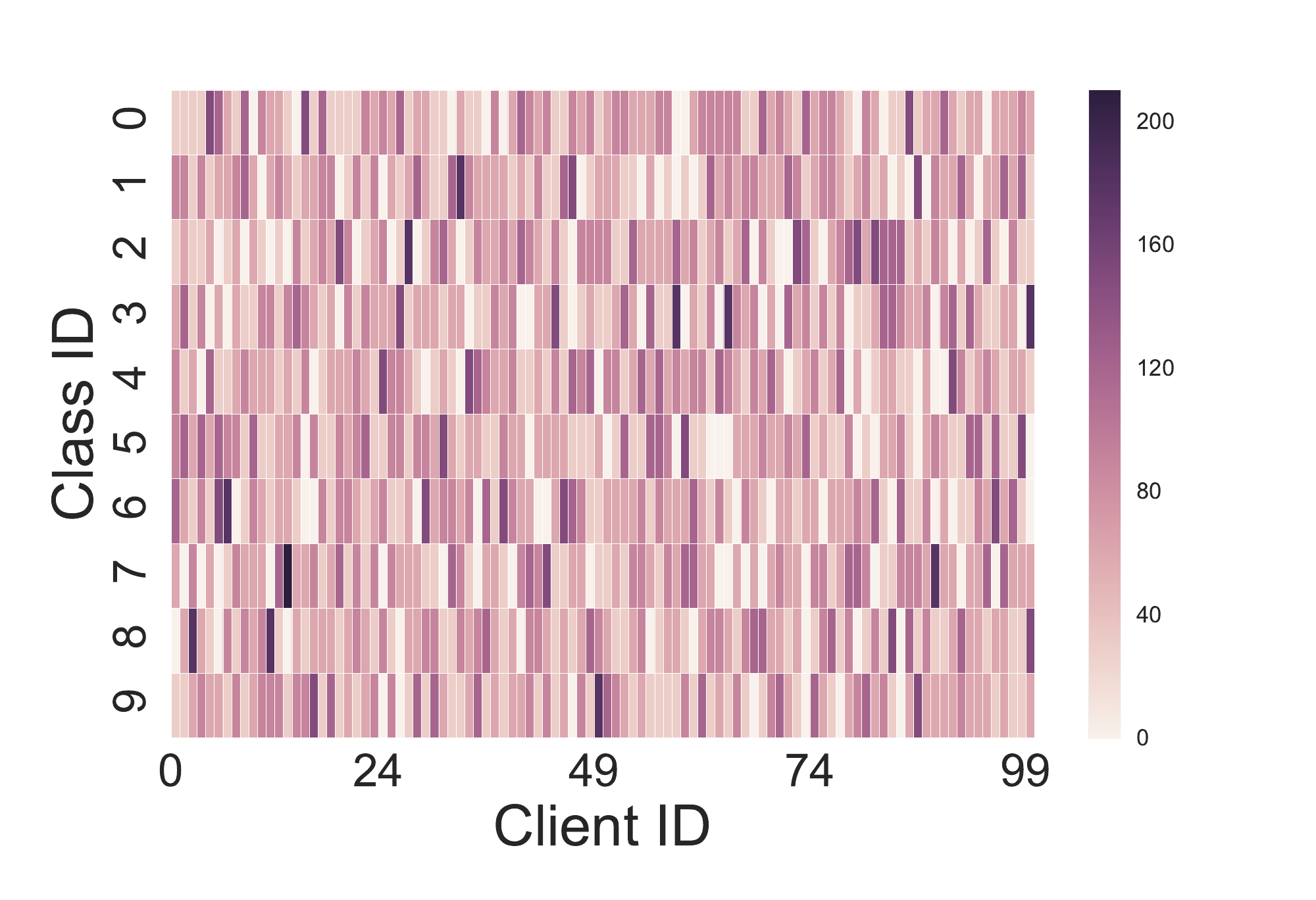}}
	\label{fig: dataset_case1}
	\hspace{-0.5cm}
	\subfigure[Case \uppercase\expandafter{\romannumeral2}]{\includegraphics[width=6cm, height=4.3cm]{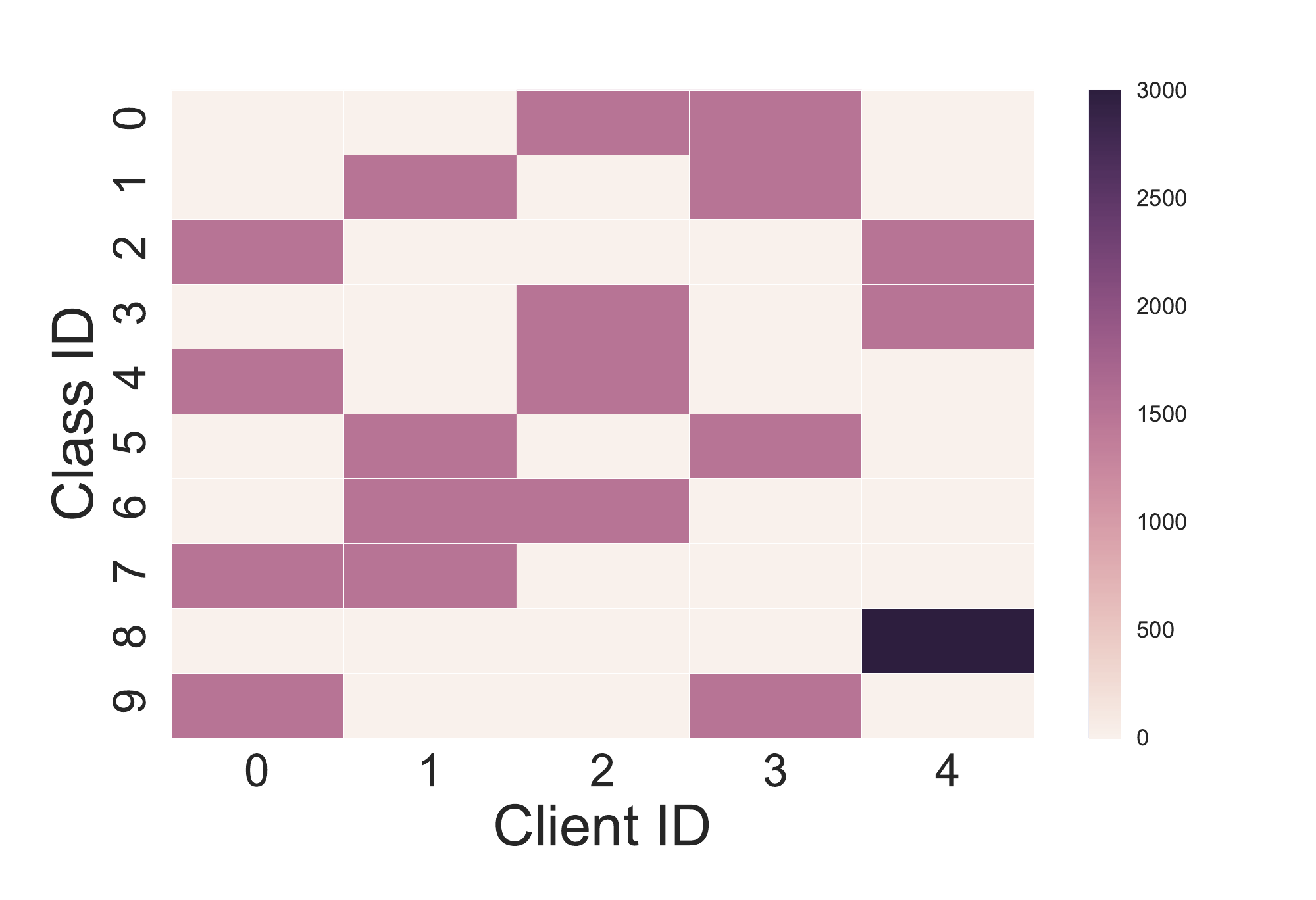}}
	\label{fig: dataset_case2}
	\hspace{-0.5cm}
	\subfigure[Case \uppercase\expandafter{\romannumeral2}]{\includegraphics[width=6cm, height=4.3cm]{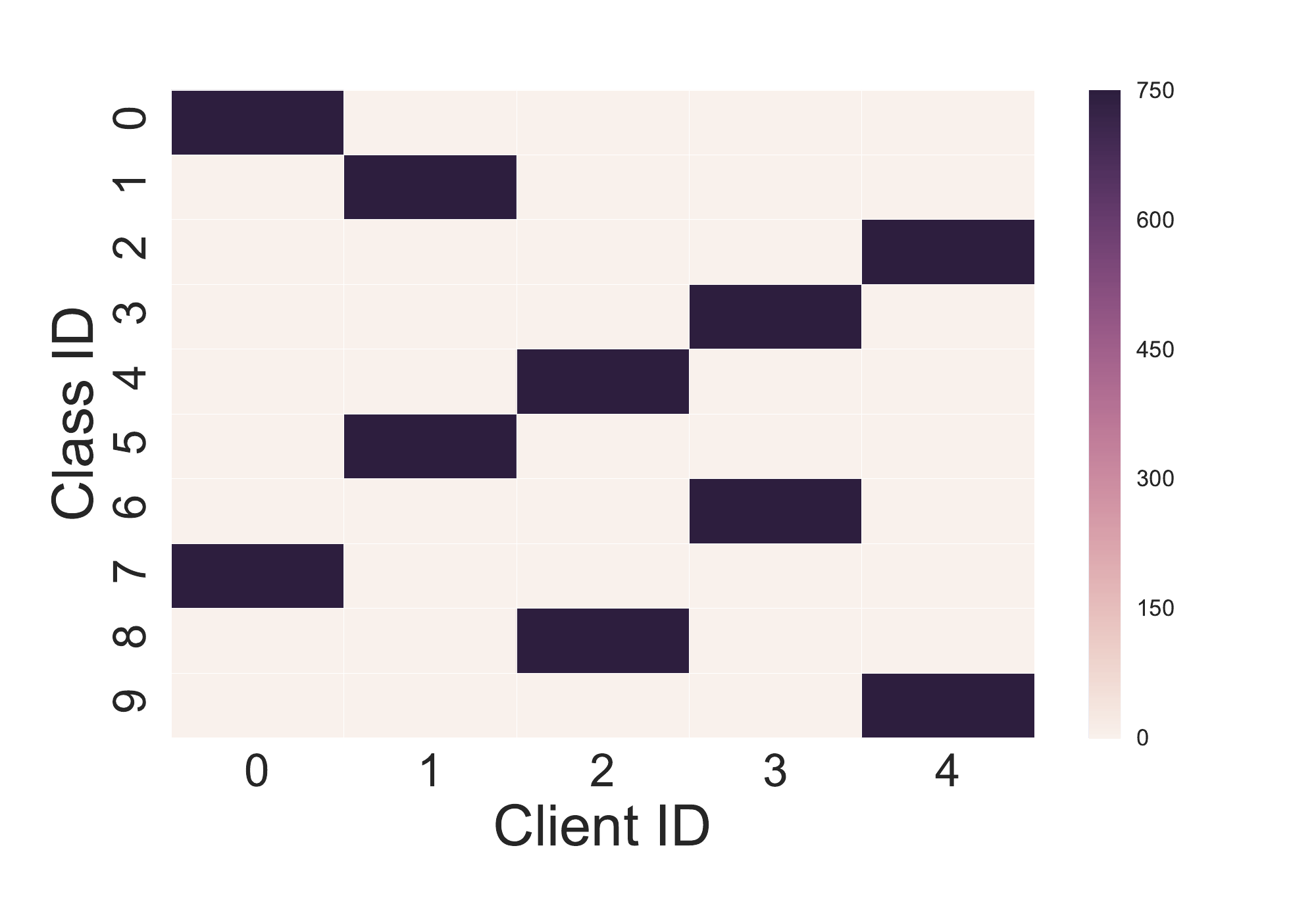}}
	\label{fig: dataset_case3}
	\caption{The data distribution of each client using non-IID data partition. The color bar denotes the number of data samples. Each rectangle represents the number of data samples of a specific class in a client.} 
	
\end{figure*}
\vspace{-0.5cm}

\section{Experiment} 
In this section, we provide the present empirical setup and results. We first describe our experimental setup (Section A), then we demonstrate the motivation for adding a fairness adjustment module based on an RL algorithm, showing that our algorithm can effectively reduce the classical federation model in the case of varying degrees of data non-uniform distribution Differences in performance on different clients while maintaining better performance. Meanwhile, we set the situation so bad that the data types between clients do not overlap at all, and compare the classification accuracy and fairness of the algorithm (Section B). Next, we compare the algorithm with the fairness goals of several baselines (Section C). Finally, we show the constraints of the algorithm (Section D).\par

\subsection{Experimental Setup}
\textbf{Federated datasets.} In this section, we will explore a suite of federated datasets based on classification tasks. These datasets include CIFAR-10 \cite{Krizhevsky2009LearningML}, CIFAR-100 \cite{Krizhevsky2009LearningML} and Fashion-MNIST \cite{Xiao2017FashionMNISTAN}. When used to compare with q-FFL and AFL, we will use a small benchmark dataset studied by \cite{Li2020FairRA} based on Fashion-MNIST.\par

\textbf{Data partitions.} In the construction of the non-IID dataset, we divide the data of each class of the N-class classification dataset into equal-sized partitions, and all clients randomly select different numbers of partitions, so that each client has local data with inconsistent quantities and categories. As shown in Fig. 3, i) Case \uppercase\expandafter{\romannumeral1}, as (a), set 100 clients, each client randomly selects 20 data blocks, take CIFAR-100 as an example; ii) Case \uppercase\expandafter{\romannumeral2}, as (b), set 5 clients, each client has 4 data blocks, take CIFAR-10 as an example, the data categories on the client overlap; iii) Case \uppercase\expandafter{\romannumeral3}, as (c), set 5 clients, each client has 2 data blocks, Taking CIFAR-10 as an example, the data categories on the client do not overlap at all.\par

\begin{figure*}
	\centering
	\subfigure[CIFAR-10]{\includegraphics[width=5.8cm, height=3cm]{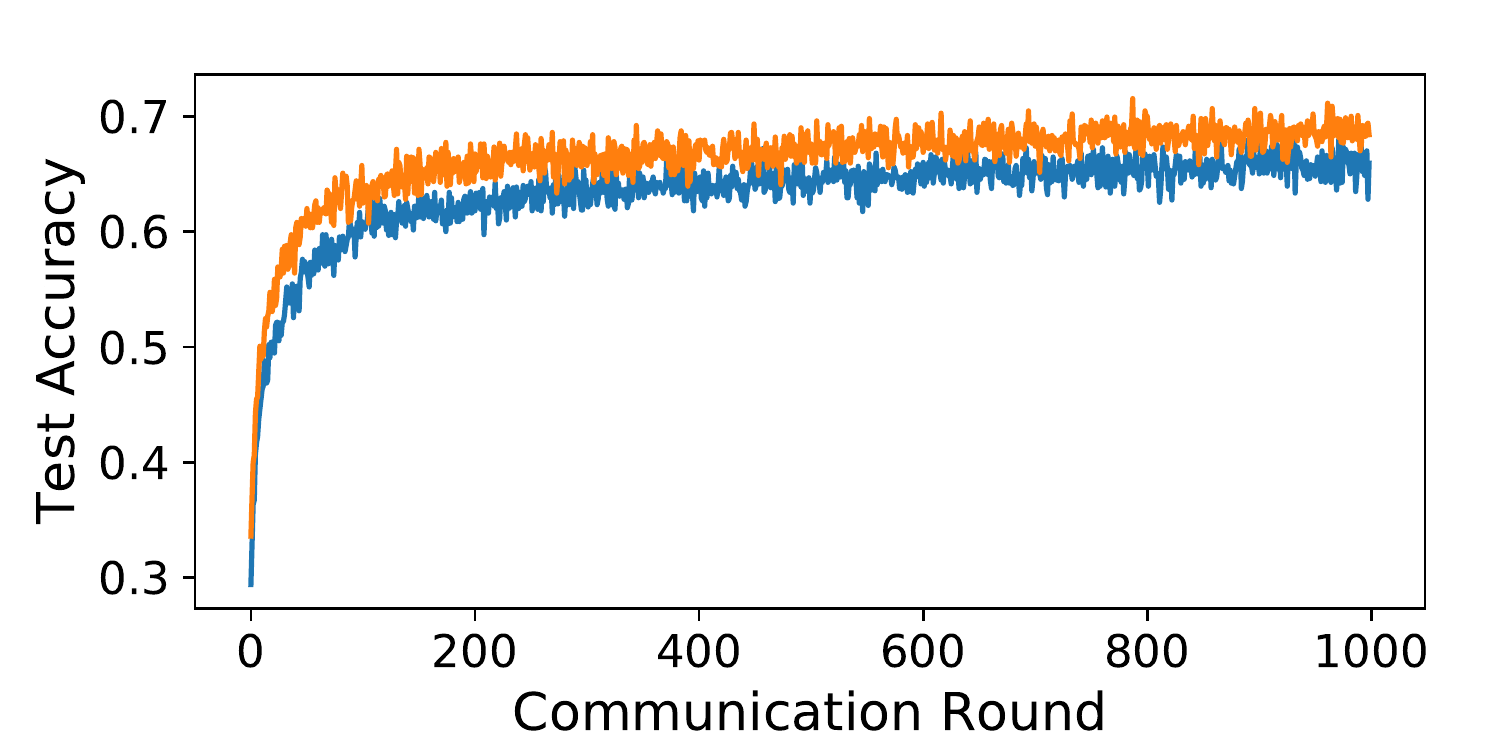}}
	\subfigure[CIFAR-100]{\includegraphics[width=5.8cm, height=3cm]{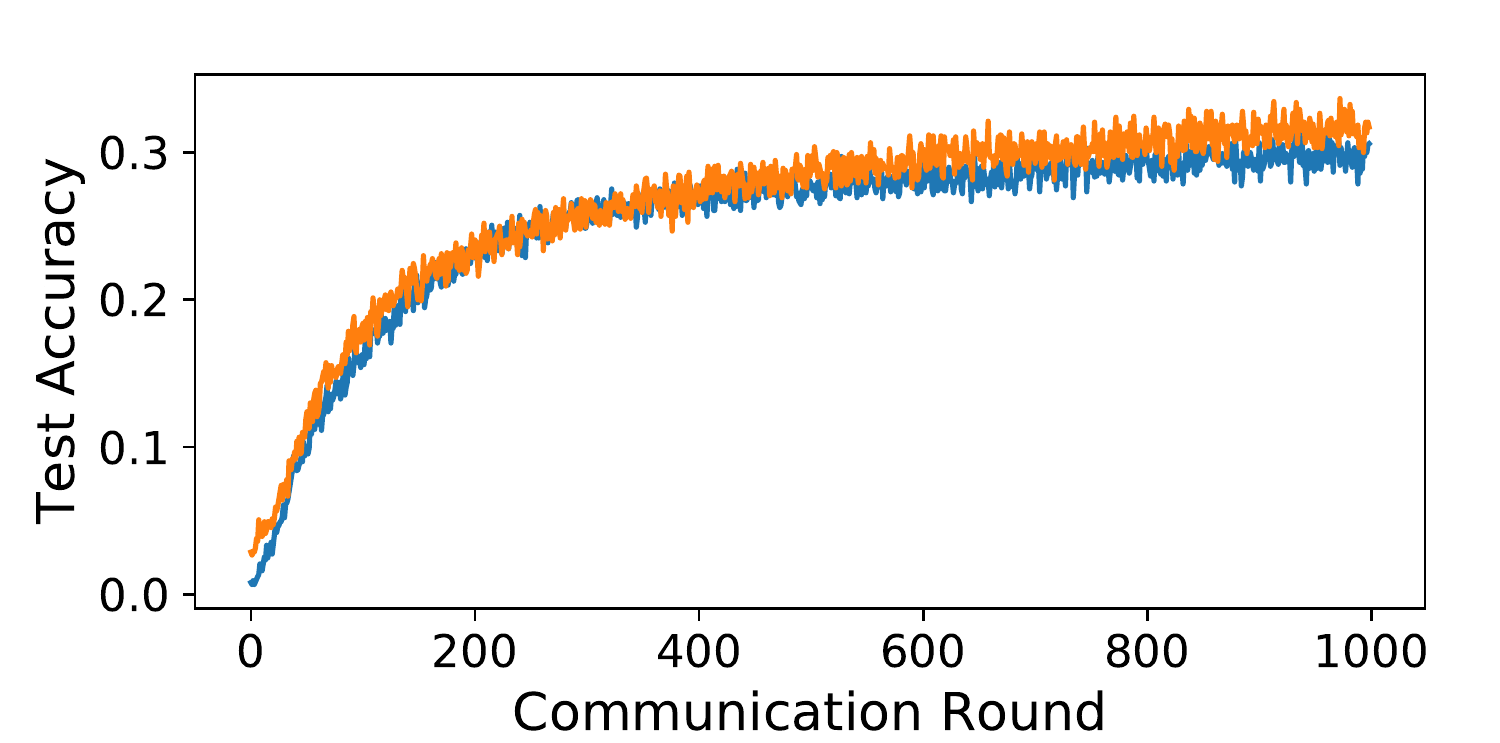}}
	\subfigure[Fashion-MNIST]{\includegraphics[width=5.8cm, height=3cm]{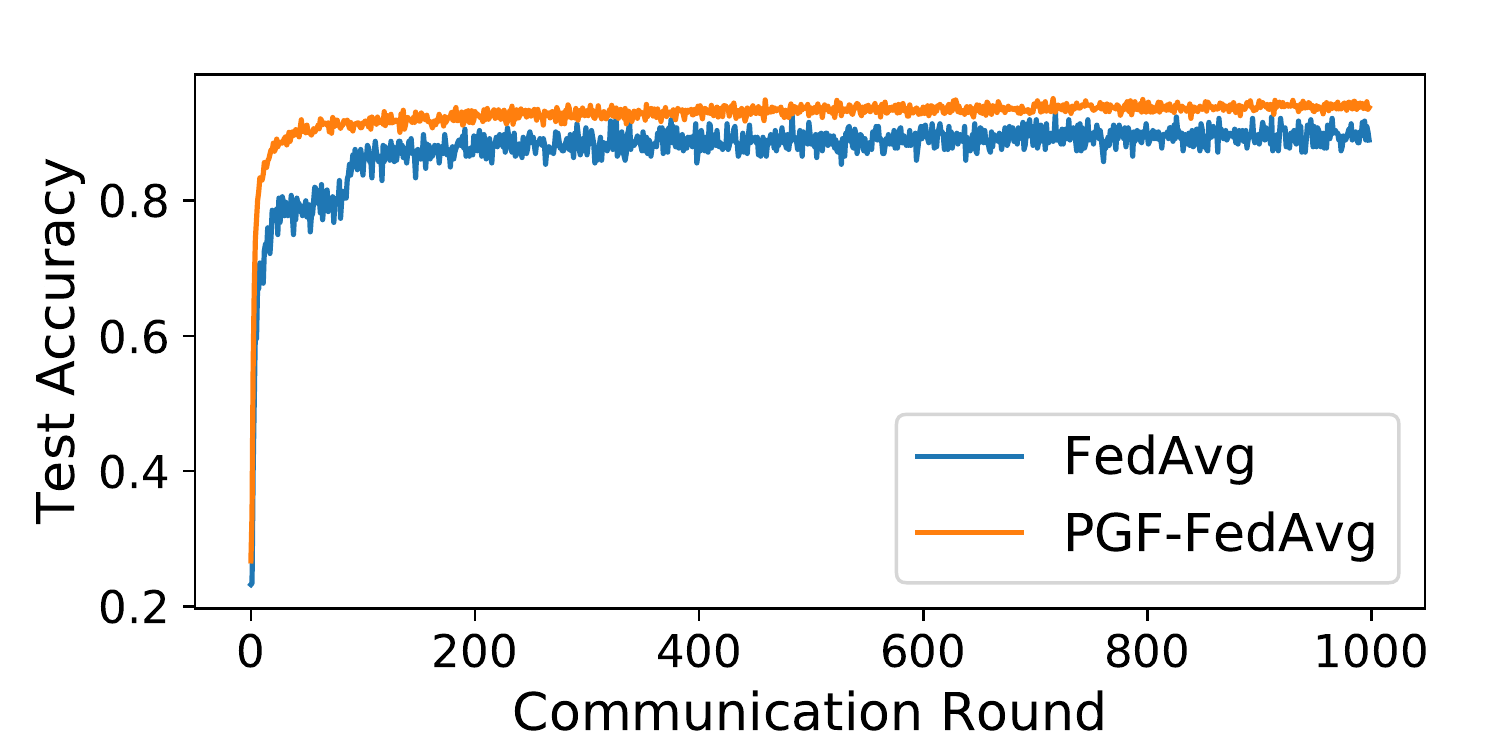}}
	\\ \vspace{-0.2cm} 
	\centering
	\subfigure[CIFAR-10]{\includegraphics[width=5.8cm, height=3cm]{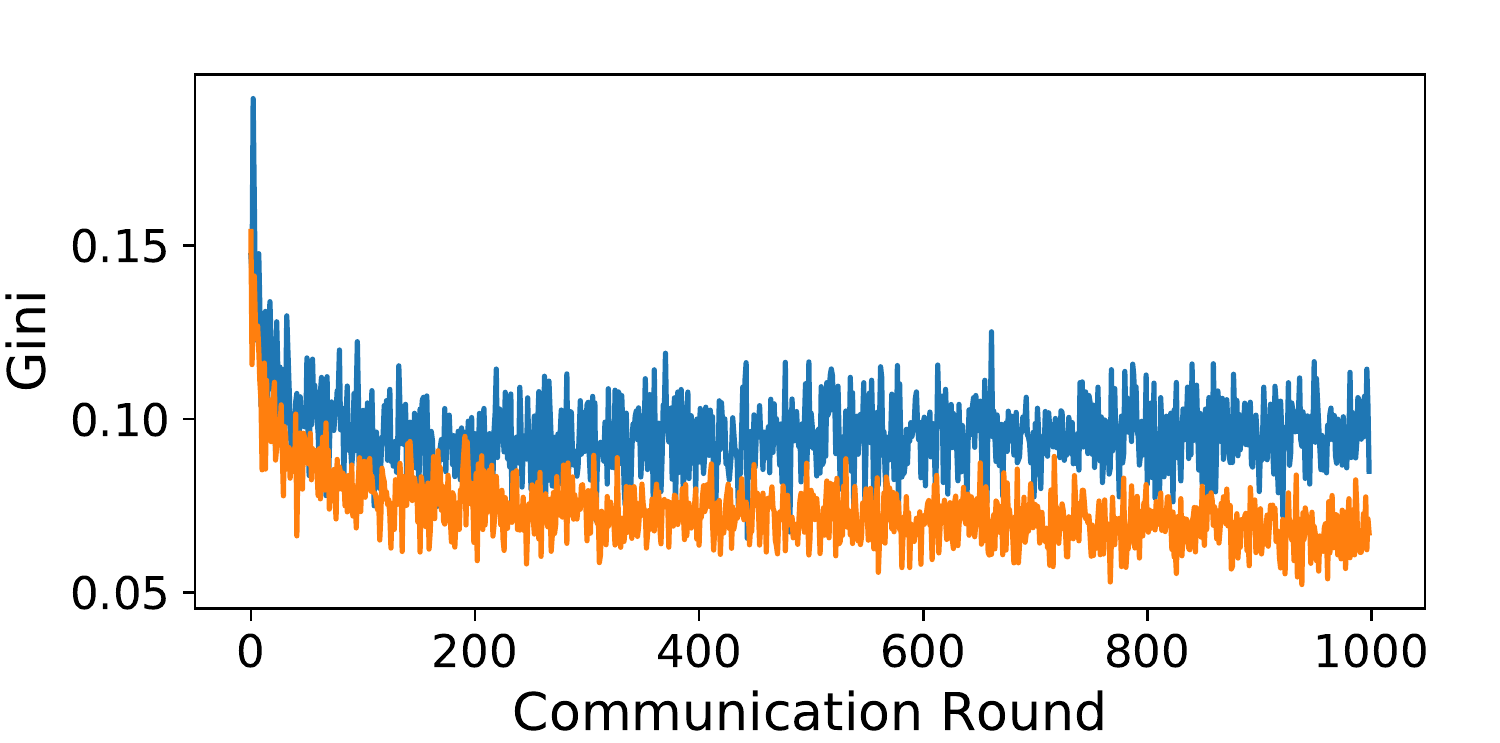}}
	\subfigure[CIFAR-100]{\includegraphics[width=5.8cm, height=3cm]{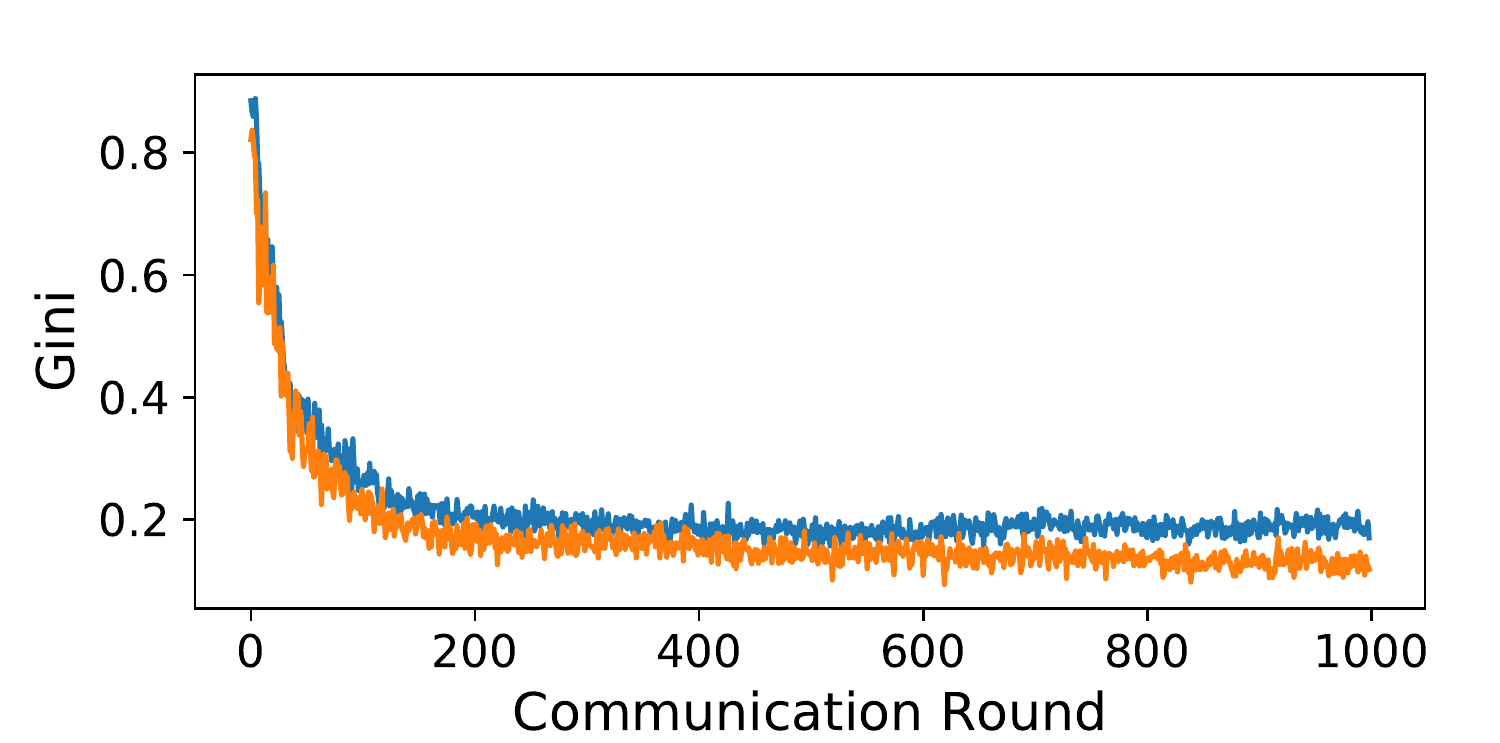}}
	\subfigure[Fashion-MNIST]{\includegraphics[width=5.8cm, height=3cm]{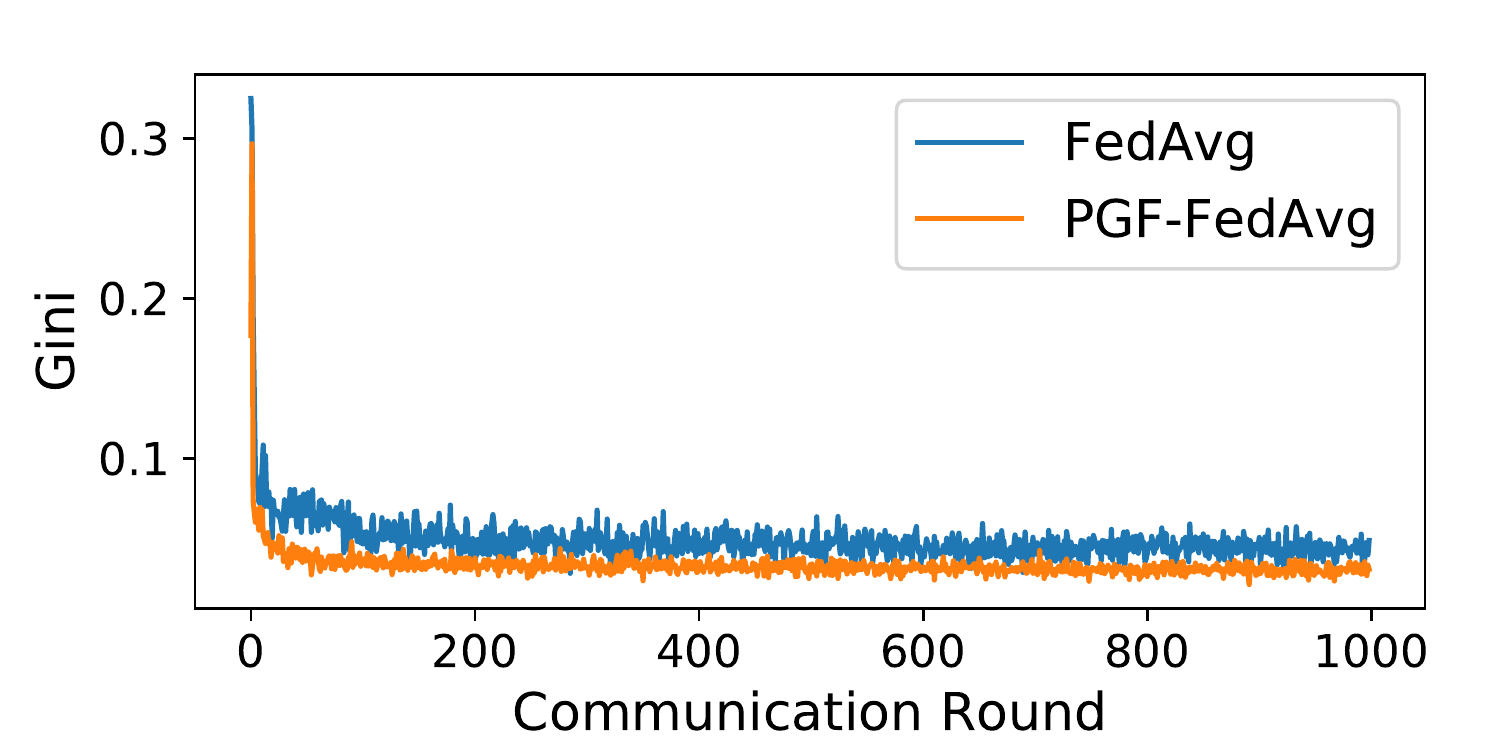}}
	\\ \vspace{-0.2cm} 
	\centering
	\subfigure[CIFAR-10]{\includegraphics[width=5.8cm, height=3cm]{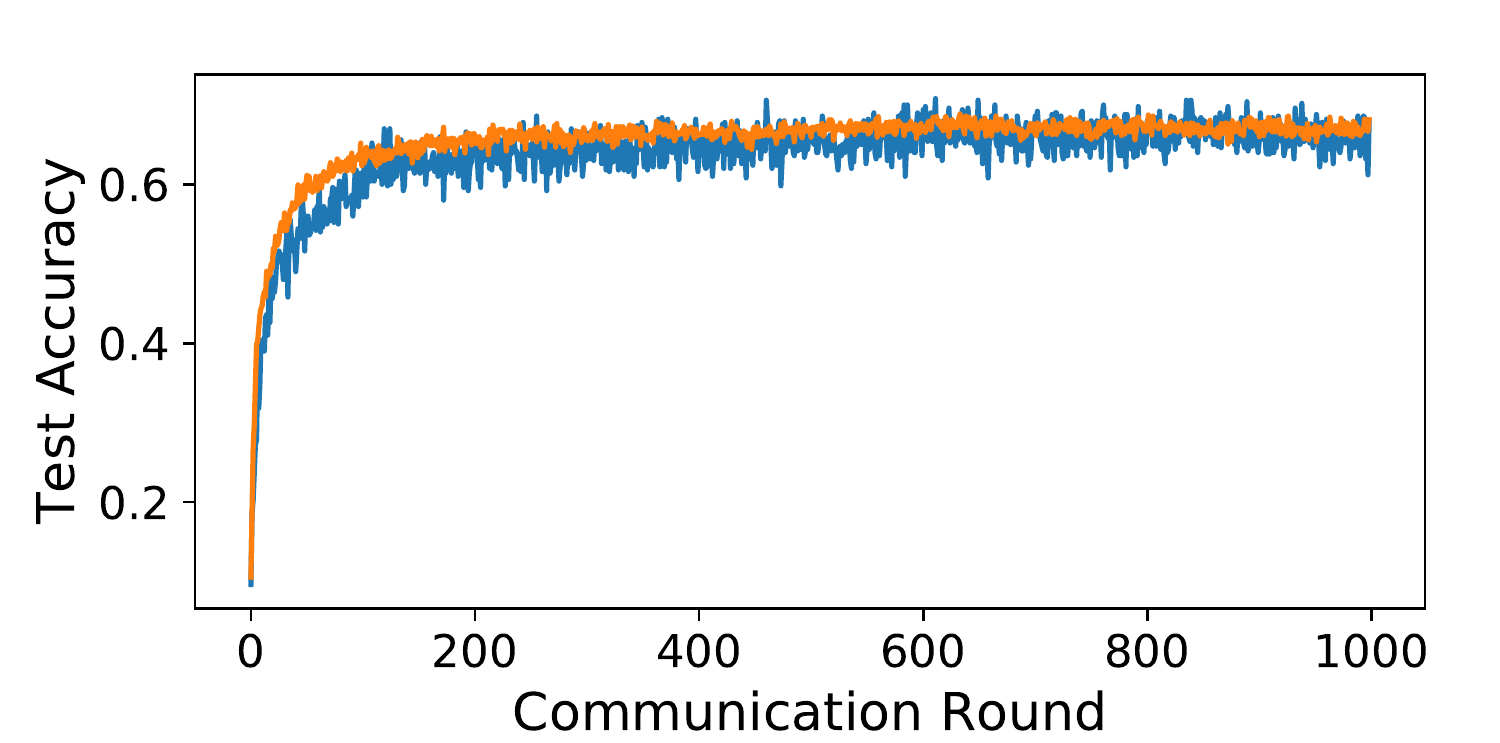}}
	\subfigure[CIFAR-100]{\includegraphics[width=5.8cm, height=3cm]{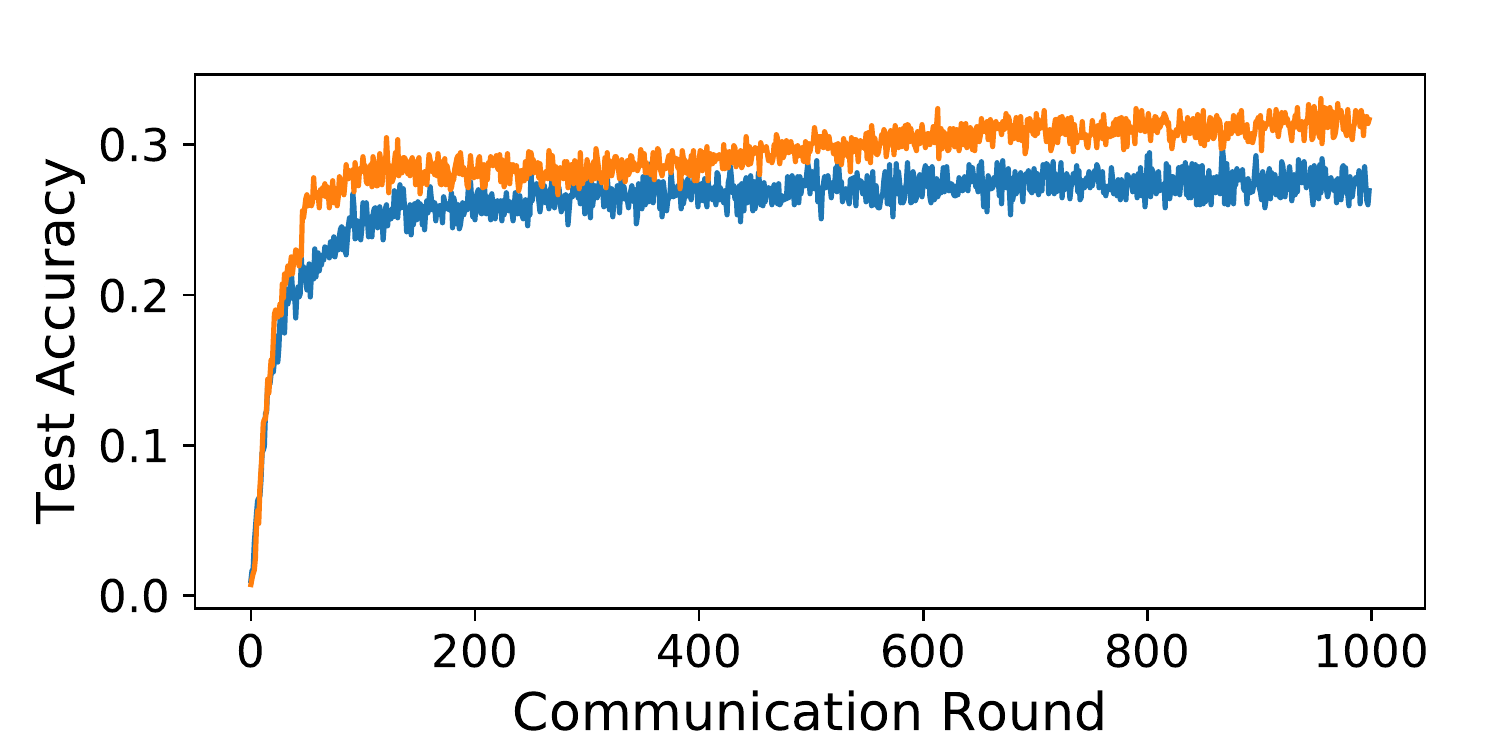}}
	\subfigure[Fashion-MNIST]{\includegraphics[width=5.8cm, height=3cm]{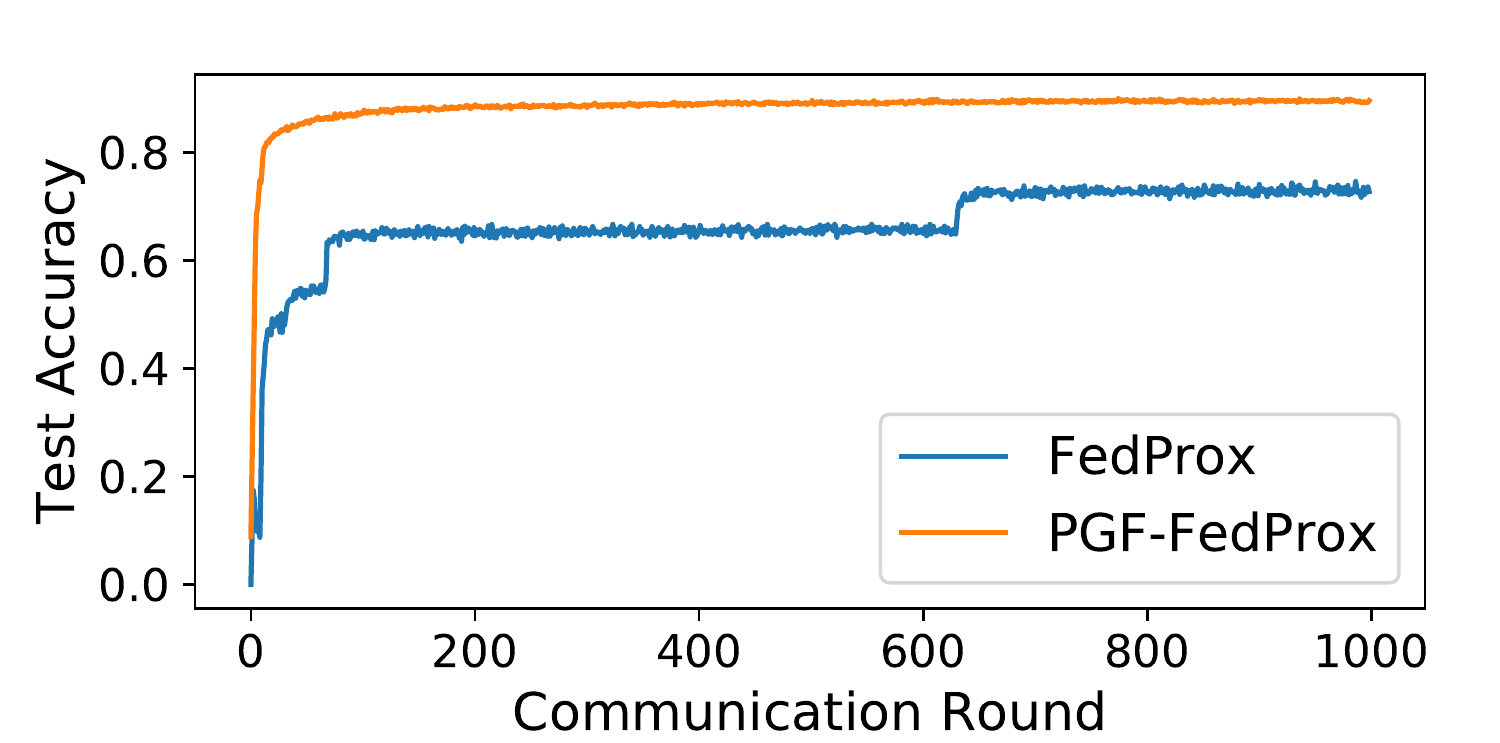}}
	\\ \vspace{-0.2cm} 
	\centering
	\subfigure[CIFAR-10]{\includegraphics[width=5.8cm, height=3cm]{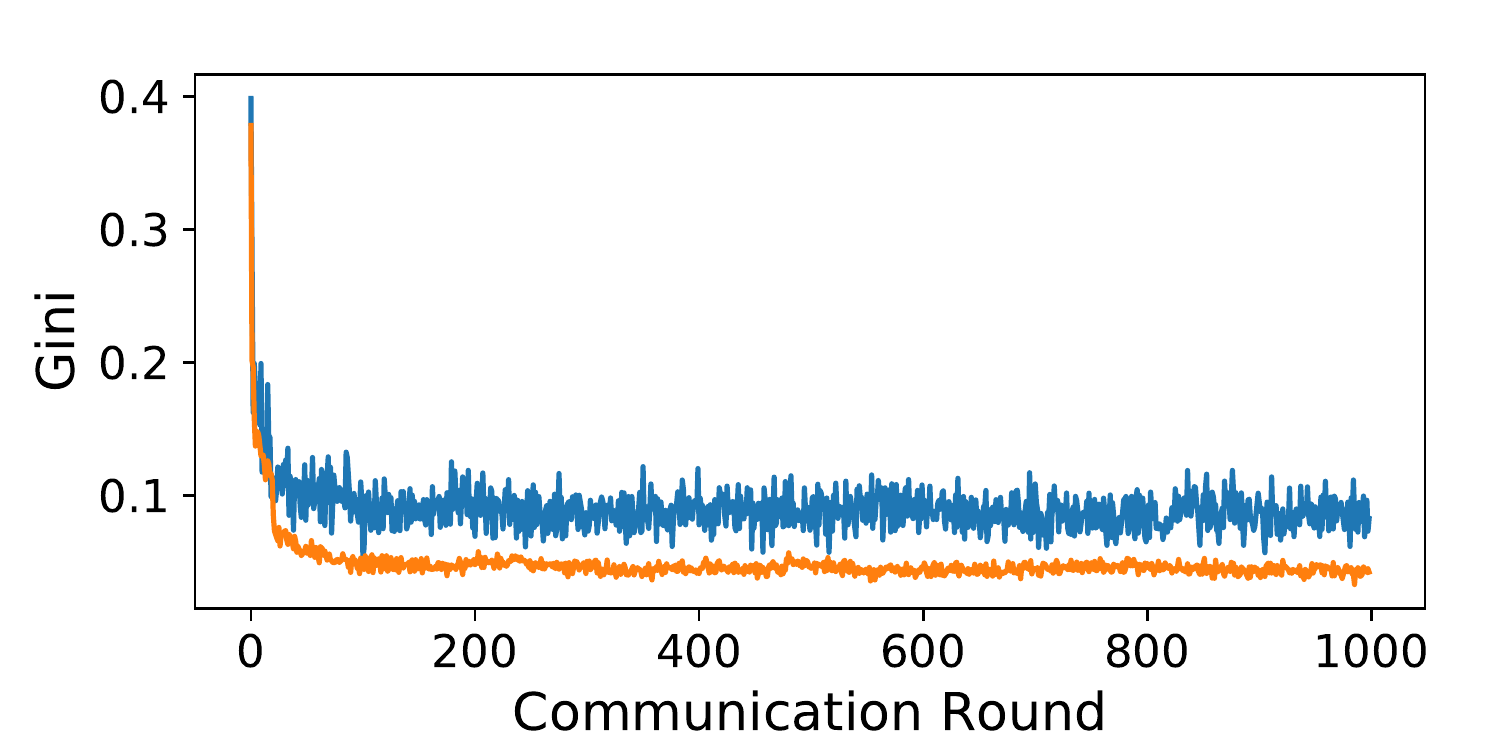}}
	\subfigure[CIFAR-100]{\includegraphics[width=5.8cm, height=3cm]{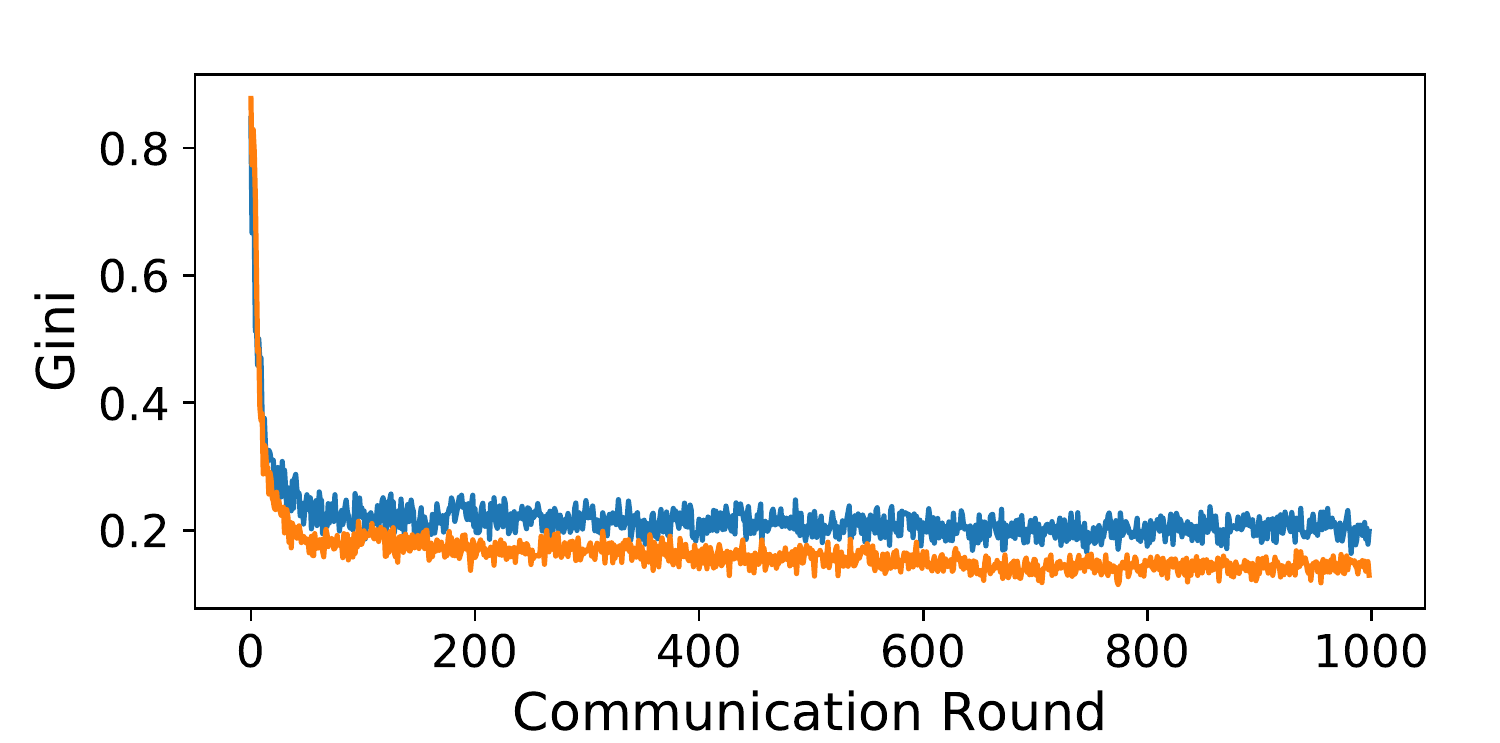}}
	\subfigure[Fashion-MNIST]{\includegraphics[width=5.8cm, height=3cm]{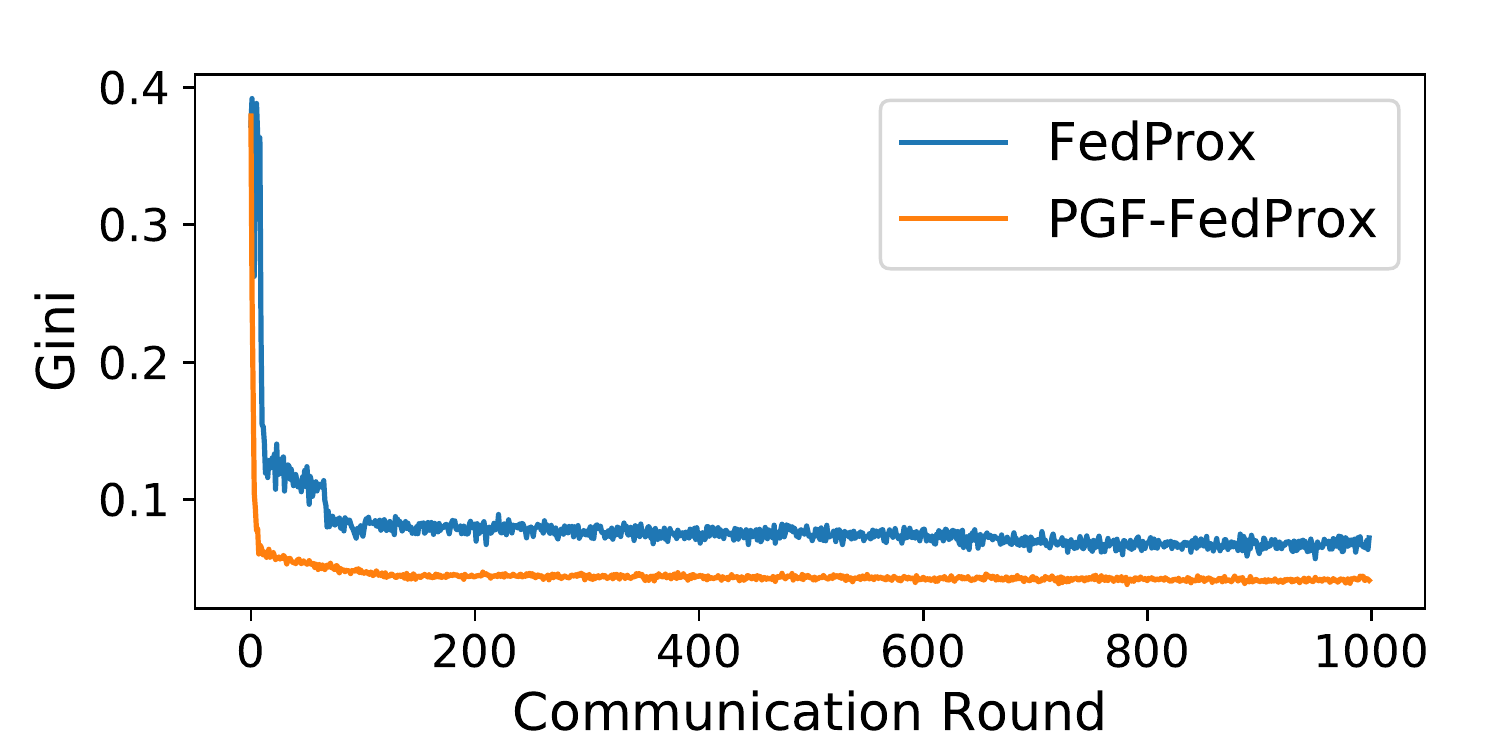}}
	\caption{From left to right are the experimental results on the CIFAR-10, CIFAR-100 and Fashion-MNIST datasets, respectively. Data distribution refer to Fig. 3, case \uppercase\expandafter{\romannumeral1}. The first and second rows show the performance (average test accuracy on clients) and fairness (definition 1) of our algorithm compared with FedAvg. The third and fourth rows show the performance and fairness of our algorithm compared with FedProx. Note that our fairness plugin has the same parameter settings as the original algorithm before adding.} 
\end{figure*}

\textbf{Implementation.} We implemented all the codes based on Pytorch, using a server and N clients to simulate a federated network, where N is the total number of clients.\par

\hspace{-0.5cm}
\subsection{Fairness of PG-FFL}
In this section, we show the efficiency of our fairness adjustment plug-in combined with FedAvg and FedProx, which are both classical and effective FL algorithms. We set up 100 clients to train on CIFAR-10, CIFAR-100 and Fashion-MNIST, respectively, and the data distribution of clients is as case \uppercase\expandafter{\romannumeral1} shown in Fig. 3(c), the local data distribution is highly heterogeneous. On the model selection, we train CIFAR-10 and CIFAR-100 by CNN, Fashion-MNIST by a four-layer MLP. The fairness adjustment plug-in based on the policy gradient reinforcement learning algorithm uses four-layer multi-layer-perceptron to learn the optimal aggregation strategy.\par

For each communication epoch, FedProx and PGF-FedProx are set to train locally for 5 epochs, while FedAvg and PGF-FedAvg execute one epoch locally. we can observe that in Fig. 4, our algorithm basically maintains the same convergence speed as the baseline, the average accuracy is slightly improved or more stable, and the fairness is significantly enhanced.\par

Next, we further exacerbate the inhomogeneity of the data distribution, verifying that our proposed PG-FFL algorithm provides a fairer solution for federated data. We will validate on a 10-class classification problem, reduce the number of clients trained on Cifar10 and Fashion MNIST to 5, and have them all participate in each round of federated updates. In Table 1, we compare the final test accuracy and fairness of our proposed fairness adjustment plugin combined with FedAvg and FedProx, respectively, for the data IID and non-IID distribution as case \uppercase\expandafter{\romannumeral2} and case \uppercase\expandafter{\romannumeral3}. Note that, when each client has only two classes, their data classes will not overlap at all.\par

We can observe that with the deterioration of the client data distribution non-IID situation, the baseline algorithm can significantly improve the fairness after adding the fairness adjustment plug-in, and also improve the average accuracy of the model on the client. When the client data distributes IID, our method sometimes can also improve the fairness of the model, but it will cause a certain loss of accuracy. We speculate that it is because the fairness and average accuracy are considered in our RL model reward at this time, so In order to ensure a high degree of fairness between the client test accuracy, it prevents our algorithm from pursuing higher average accuracy, thus causing some accuracy loss.\par

\begin{table}[ht]
	\caption{The accuracy and fairness comparison of PG-FFL and baselines tested on datasets with varying degrees of non-IID.}
	\setlength{\tabcolsep}{1.2mm}{
		\renewcommand\arraystretch{1.5}
		\begin{tabular}{ccccccc}
			\multicolumn{7}{c}{CIFAR-10}                                                                                                                           \\ \hline
			\multicolumn{1}{c}{Non-IID level} & \multicolumn{2}{c}{IID}         & \multicolumn{2}{c}{Case 2} & \multicolumn{2}{c}{Case 3} \\
			\multicolumn{1}{c}{}              & acc(↑)       & Gini(↓)           & acc(↑)            & Gini(↓)              & acc(↑)           & Gini(↓)             \\ \hline
			\multicolumn{1}{c}{FedAvg}        & 0.675          & 0.098          & 0.612               & 0.102             & 0.454              & 0.161            \\
			\multicolumn{1}{c}{PGF-FedAvg}    & \textbf{0.683} & \textbf{0.042} & \textbf{0.665}      & \textbf{0.049}    & \textbf{0.488}     & \textbf{0.053}   \\
			\multicolumn{1}{c}{FedProx}       & \textbf{0.708} & 0.107          & 0.595               & 0.110             & 0.422              & 0.154            \\
			\multicolumn{1}{c}{PGF-FedProx}   & 0.689          & \textbf{0.073} & \textbf{0.608}      & \textbf{0.037}    & \textbf{0.453}     & \textbf{0.031}   \\ \hline
			
			\multicolumn{7}{c}{CIFAR-100}                                                                                                                          \\ \hline
			\multicolumn{1}{c}{Non-IID level} & \multicolumn{2}{c}{IID}         & \multicolumn{2}{c}{Case 2} & \multicolumn{2}{c}{Case 3} \\
			\multicolumn{1}{c}{}              & acc(↑)       & Gini(↓)           & acc(↑)            & Gini(↓)              & acc(↑)           & Gini(↓)             \\ \hline
			\multicolumn{1}{c}{FedAvg}        & 0.493          & 0.130          & 0.468               & 0.133             & 0.316              & 0.151            \\
			\multicolumn{1}{c}{PGF-FedAvg}    & \textbf{0.502} & \textbf{0.093} & \textbf{0.491}      & \textbf{0.068}    & \textbf{0.337}     & \textbf{0.093}   \\
			\multicolumn{1}{c}{FedProx}       & 0.509          & 0.133          & 0.467               & 0.143             & 0.299              & 0.164            \\
			\multicolumn{1}{c}{PGF-FedProx}   & \textbf{0.514} & \textbf{0.084} & \textbf{0.483}      & \textbf{0.079}    & \textbf{0.331}     & \textbf{0.085}   \\ \hline
			
			\multicolumn{7}{c}{Fashion-MNIST}                                                                                                                      \\ \hline
			\multicolumn{1}{c}{Non-IID level} & \multicolumn{2}{c}{IID}         & \multicolumn{2}{c}{ Case 2} & \multicolumn{2}{c}{Case 3} \\
			\multicolumn{1}{c}{}              & acc(↑)       & Gini(↓)           & acc(↑)            & Gini(↓)              & acc(↑)           & Gini(↓)             \\ \hline
			\multicolumn{1}{c}{FedAvg}        & 0.869          & 0.032          & 0.850               & 0.063             & 0.737              & 0.074            \\
			\multicolumn{1}{c}{PGF-FedAvg}    & \textbf{0.884} & \textbf{0.021} & \textbf{0.874}      & \textbf{0.024}    & \textbf{0.796}     & \textbf{0.033}   \\
			\multicolumn{1}{c}{FedProx}       & 0.879          & 0.021          & 0.851               & 0.045             & 0.828              & 0.037            \\
			\multicolumn{1}{c}{PGF-FedProx}   & \textbf{0.883} & \textbf{0.017} & \textbf{0.854}      & \textbf{0.031}    & \textbf{0.836}     & \textbf{0.020}   \\ \hline
		\end{tabular}}
\end{table}

\subsection{Comparison With Other Fair Federated Learning Algorithms}
Next, we compare with other two algorithms that also aim to address fairness issues in federated networks.\par

In the experiments in this section, we implement a very extreme case where each client has only a completely disjoint class of data. Using the same experimental setup as \cite{Mohri2019AgnosticFL}: The Fashion-MNIST dataset \cite{Xiao2017FashionMNISTAN} is an MNIST-like dataset where images are classified into 10 categories of clothing instead of handwritten digits. We extract a subset of the data labeled with three categories - shirts/tops, pullovers, and shirts, and divide this subset into three clients, each containing a category of clothing. We then train a classifier for these three classes using logistic regression and the Adam optimizer. Since the clients here uniquely identify the labels, in this experiment we did not compare with models trained on specific baselines.\par

We observe in Table 2 that our algorithm performs better both in the final average accuracy and fairness between clients.\par

\vspace{-0.3cm} 
\begin{table}[ht]
	\caption{The accuracy and fairness comparison with PG-FFL and other fairness algorithms in the case of extreme data non-IID distribution.}
	\centering
	\renewcommand\arraystretch{1.5}
	\begin{tabular}{lccccc}
		\hline
		& \multicolumn{2}{c}{All Clients} & Shirts   & Pullovers     & T-shirts        \\
		& acc(↑)       & Gini(↓)           & acc(↑)           & acc(↑)           & acc(↑)           \\ \hline
		q-FFL(q=0) & 78.8           & 0.084          & 66.0          & \textbf{84.5} & \textbf{85.9} \\
		AFL   & 78.2           & 0.046          & 71.4          & 81.0          & 82.1          \\
		PGF-FedAvg  & \textbf{79.1}  & \textbf{0.027} & \textbf{74.2} & 80.5          & 82.6          \\ \hline
	\end{tabular}
\end{table}
\vspace{-0.5cm} 
\subsection{Scalability Analysis}

In this section, we continue the experimental setup in section B by combining our proposed fairness adjustment plugin with FedAvg and FedProx, respectively, to modify the percentage of participating updates in each round, but keep the total number of clients unchanged. It can be observed from Fig. 5 that there is an upper limit on the scalability of the algorithm, and the greater the proportion of clients participating in the global update, the better the effect. We guess it is because each time the RL algorithm will output the proportion of the client participating in the update, but when the proportion of the client participating in the update is small, the non-participating clients cannot obtain the updated aggregate participation weight in time, which will affect the experimental results. We use STD as the fairness indicator here because that is consistent with the test accuracy dimension and is more likely to show volatility. It's easy to see that PG-FFL can still improve the fairness of the federated network under other fairness definition.\par

\begin{figure}[ht]
	\centering
	\subfigure[10\% from 100 clients]{\includegraphics[width=4cm]{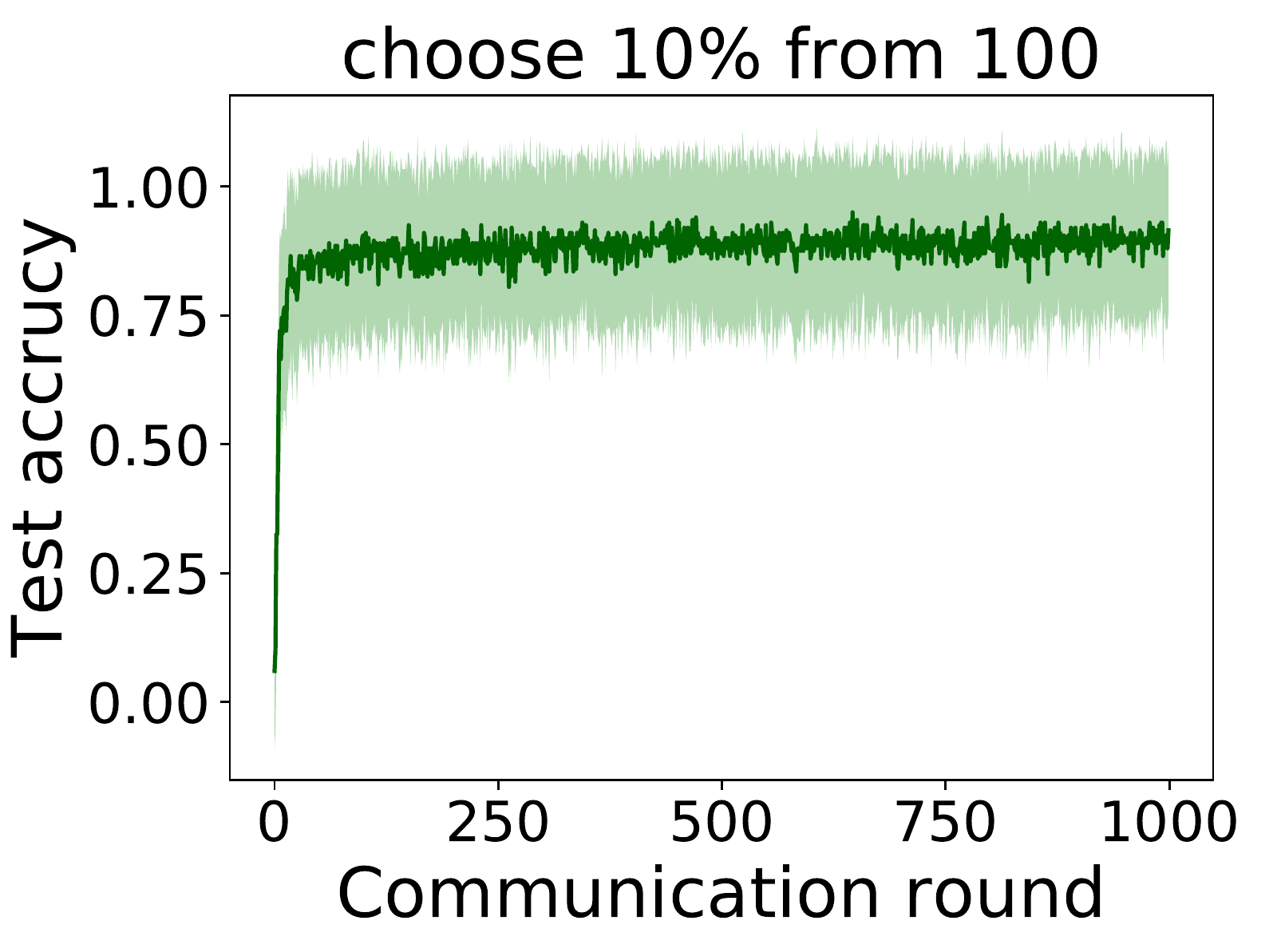}}
	\subfigure[25\% from 100 clients]{\includegraphics[width=4cm]{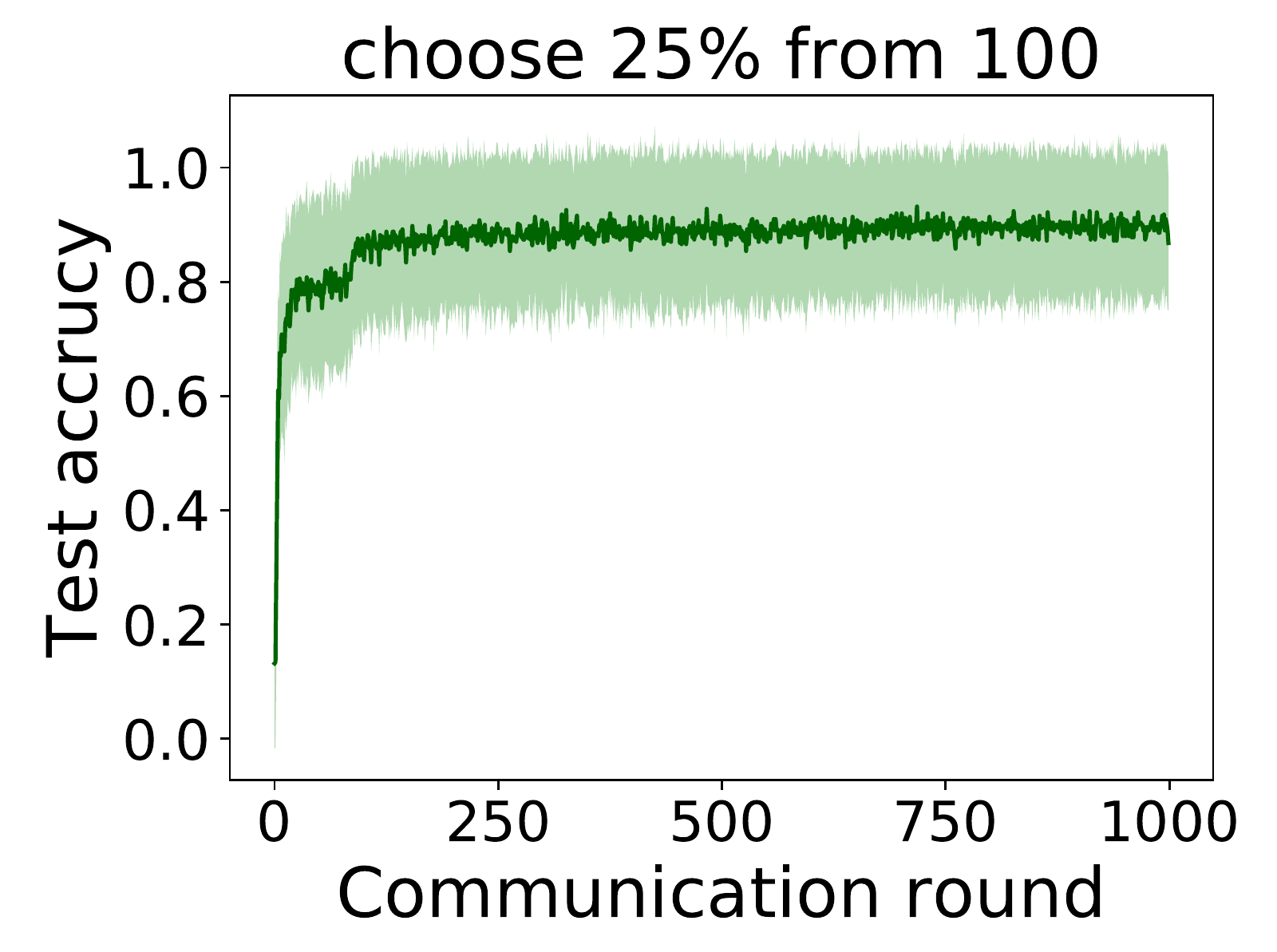}}
	\\ 
	\centering
	\subfigure[50\% from 100 clients]{\includegraphics[width=4cm]{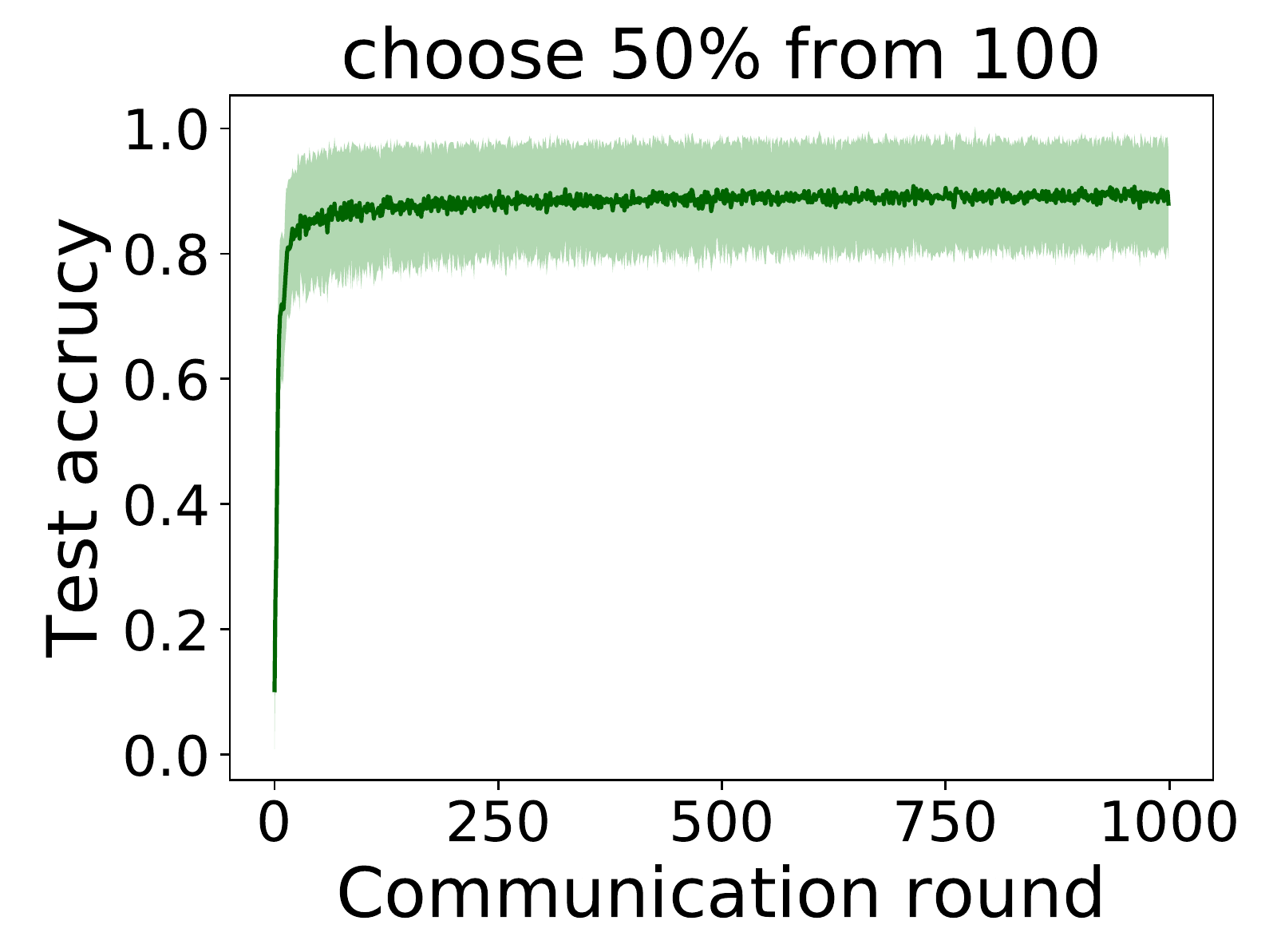}}
	\subfigure[75\% from 100 clients]{\includegraphics[width=4cm]{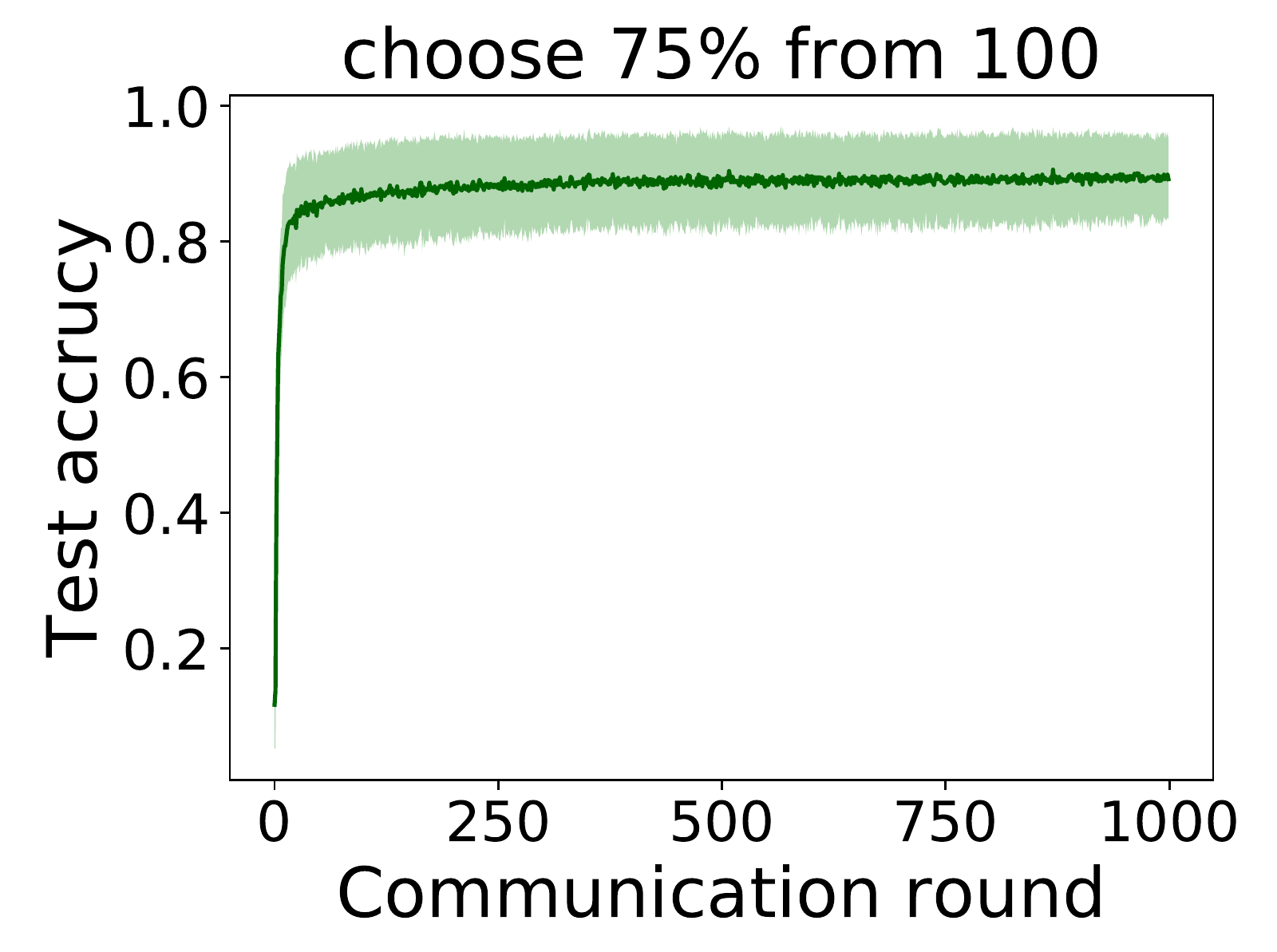}}
	\caption{The solid line is the client's average validation accuracy, the green range is the accuracy STD on different clients, and the smaller the area is, the fairer it is.} 
\end{figure}

\section{Conclusion}
In this paper, we propose fairness as a new optimization objective defined by the Gini coefficient of clients' validation accuracy, which is based on realistic considerations that encourage fairer accuracy distribution across clients in federated learning. We design a reinforcement learning plug-in to apply federated algorithms to solve this problem in large-scale networks, and experiments demonstrate that PG-FFL can be regarded as a fairness add-on for any global objective. We illustrate the fairness and superiority of PG-FFL on a set of federated datasets, and experimental results show that our framework outperforms baseline methods in terms of overall performance, fairness, and convergence speed.\par

 \section*{Acknowledgement}
This paper is supported by the Key Research and Development Program of Guangdong Province under grant No. 2021B0101400003 and Shenzhen Basic Research Program (Natural Science Foundation) Key Program of Fundamental Research (No. JCYJ20200109143016563). Corresponding authors are Jianzong Wang from Ping An Technology (Shenzhen) Co., Ltd (jzwang@188.com) and Yuhan Dong from Tsinghua University (dongyuhan@sz.tsinghua.edu.cn).

\footnotesize
\bibliography{refs}
\bibliographystyle{IEEEbib}

\end{document}